\documentclass[preprint,review,12pt]{elsarticle}

\usepackage{amssymb}

\usepackage{amsmath,amssymb,amsfonts}
\usepackage{algorithmic}
\usepackage{graphicx}
\usepackage{ulem}
\usepackage{soul}
\usepackage{textcomp}
\usepackage{xcolor}
\usepackage{float}
\usepackage{overpic}
\usepackage{multirow}
\usepackage{subfigure}
\usepackage{booktabs}

\usepackage[colorlinks,
linkcolor=red,
anchorcolor=red,
citecolor=green]{hyperref}
\def\BibTeX{{\rm B\kern-.05em{\sc i\kern-.025em b}\kern-.08em
		T\kern-.1667em\lower.7ex\hbox{E}\kern-.125emX}}

\journal{Expert Systems with Applications}

\begin{document}

\title{Orientation and Context Entangled Network for Retinal Vessel Segmentation}

\author[1,2,4]{Xinxu Wei}
\ead{xinxu.wei@mail.mcgill.ca}

\author[1]{Kaifu Yang}
\ead{yangkf@uestc.edu.cn}

\author[2,3,4]{Danilo Bzdok}
\ead{danilobzdok@gmail.com}
\author[1]{Yongjie Li*}
\ead{liyj@uestc.edu.cn}


\cortext[cor1]{Corresponding author}

\address[1]{MOE Key Lab for Neuroinformation, School of Life Sciences and Technology, University of Electronic Science and Technology of China}

\address[2]{McConnell Brain Imaging Centre, Montreal Neurological Institute (MNI), McGill University, Montreal, QC, Canada}

\address[3]{Department of Biomedical Engineering, Faculty of Medicine, McGill University, Montreal, QC, Canada}

\address[4]{Mila - Quebec Artificial Intelligence Institute, Montreal, QC, Canada}

\begin{abstract}
Most of the existing deep learning based methods for vessel segmentation neglect two important aspects of retinal vessels, one is the orientation information of vessels, and the other is the contextual information of the whole fundus region. In this paper, we propose a robust Orientation and Context Entangled Network (denoted as OCE-Net), which has the capability of extracting complex orientation and context information of the blood vessels. To achieve complex orientation aware, a Dynamic Complex Orientation Aware Convolution (DCOA Conv) is proposed to extract complex vessels with multiple orientations for improving the vessel continuity. To simultaneously capture the global context information and emphasize the important local information, a Global and Local Fusion Module (GLFM) is developed to simultaneously model the long-range dependency of vessels and focus sufficient attention on local thin vessels. A novel Orientation and Context Entangled Non-local (OCE-NL) module is proposed to entangle the orientation and context information together. In addition, an Unbalanced Attention Refining Module (UARM) is proposed to deal with the unbalanced pixel numbers of background, thick and thin vessels.
Extensive experiments were performed on several commonly used datasets (DRIVE, STARE and CHASEDB1) and some more challenging datasets (AV-WIDE, UoA-DR, RFMiD and UK Biobank). The ablation study shows that the proposed method achieves promising performance on maintaining the continuity of thin vessels and the comparative experiments demonstrate that our OCE-Net can achieve state-of-the-art performance on retinal vessel segmentation.

\end{abstract}


\maketitle

\section{Introduction}
\label{sec:introduction}
Retinal vessel segmentation is very helpful for ophthalmologists to diagnose eye-related diseases such as glaucoma, hypertension, diabetic retinopathy (DR) and arteriosclerosis \cite{li2021applications}, because changes of the vessel morphology and structure can indicate some symptoms of pathology. 
However, manual segmentation of retinal blood vessels is a time-consuming and laborious process because the structure of vessels is extremely complicated. In addition, manual labelling is also error-prone because there are numerous capillaries throughout the whole fundus images, which are very narrow in width and have low local contrast against to the fundus background, and these thin vessels are easily mislabeled or missed during the process of manual annotation. Therefore, automated fundus vessel segmentation \cite{mendonca2006segmentation} \cite{liskowski2016segmenting} \cite{maninis2016deep} is very meaningful and necessary for ophthalmologists to achieve a more accurate and rapid diagnosis of ophthalmic diseases.

However, vessel segmentation of the fundus images is a very challenging task. The blood vessels themselves have complex geometric structures, and arteries and veins usually have different widths. In the whole vascular system, different vascular branches have different orientations, and the capillaries are usually very small and it is difficult to seperate thin vessels from the fundus background, which is easy to be ignored. In addition, the fundus images also contain various lesion areas, and the characteristics of some lesions are very similar with blood vessels, so these vessel-like lesions are easily wrongly segmented as blood vessels.

In order to overcome these challenging problems, many traditional vessel segmentation methods \cite{chaudhuri1989detection} \cite{orlando2014learning} \cite{li2015cross} have been proposed and achieved good results. These traditional methods usually use some manually designed filters \cite{chaudhuri1989detection} \cite{soares2006retinal} to extract vascular features, and some machine learning-based methods use classifiers \cite{orlando2016discriminatively} to classify each pixel of the fundus image.

In recent years, deep learning \cite{simonyan2014very} \cite{he2016deep} has been widely used in the area of medical image processing \cite{dos2018convolutional} \cite{badar2020application}, and has achieved great success. 
Particularly, many efficient segmentation networks have been proposed \cite{maninis2016deep} \cite{li2018h} \cite{jin2019dunet} and the most well known is UNet \cite{ronneberger2015u} architecture. UNet is widely used in medical image segmentation, such as vascular segmentation \cite{jin2019dunet}, lesion area segmentation\cite{kou2020enhanced}, organ and tissue segmentation\cite{fu2018joint}.

To improve the accuracy of vascular segmentation, most of previous deep learning-based fundus vessel segmentation methods attempt to increase the depth and width of the networks, and expand the receptive field by stacking numerous local convolution kernels. However, most of these methods pay little attention to two important information of fundus vessels: the orientation and the context information, which are great important for accurate vessel segmentation. 

As for the orientation information, unlike the instance segmentation in natural images, the blood vessels in fundus images have extremely complex orientation information due to numerous furcations and branches. Capturing this complex orientation information is very helpful to improve the accuracy of vessel segmentation and the continuity of thin vessels. As for the context information, in the whole vasculature, the overall skeleton of blood vessels presents certain distribution patterns, for example, the symmetry and relative position, orientation and shape of each blood vessel branch relative to the whole vascular skeleton. In addition, some vessels are occluded by lesions, which makes them difficult to be detected.
Capturing global context information can allow the network to learn the distribution of blood vessels from a holistic perspective, which can alleviate the problem of occlusion.
This is the motivation of this work. 

We highlight the contributions of this work as follow:

\begin{itemize}
	
	\item An Orientation and Context Entangled Network (OCE-Net) is proposed for retinal vessel segmentation by simultaneously capturing the orientation and context information of blood vessels.
	
	\item A Dynamic Complex Orientation Aware Convolution (DCOA Conv) is proposed to capture complex vessels with multiple orientations.
	
	\item A novel Global and Local Fusion Module (GLFM) is proposed to simultaneously modelling the global long-range dependencies and focusing attention on local thin vessels.
	
	\item An Orientation and Context Entangled Non-local (OCE-DNL) block is proposed to entangle the orientation and context information by introducing correlation into vanilla Non-local operation.
	
	\item An Unbalanced Attention Refining Module (UARM) is designed to refine the output features and cope with the unbalanced problem among the background, thick and thin vessels.
	
	\item Extensive experiments on multiple widely used datasets show that our method outperforms many other recent methods and achieves state-of-the-art performance. 
	
\end{itemize}

\begin{figure*}[htbp]
	\centering
	\includegraphics[width=14cm]{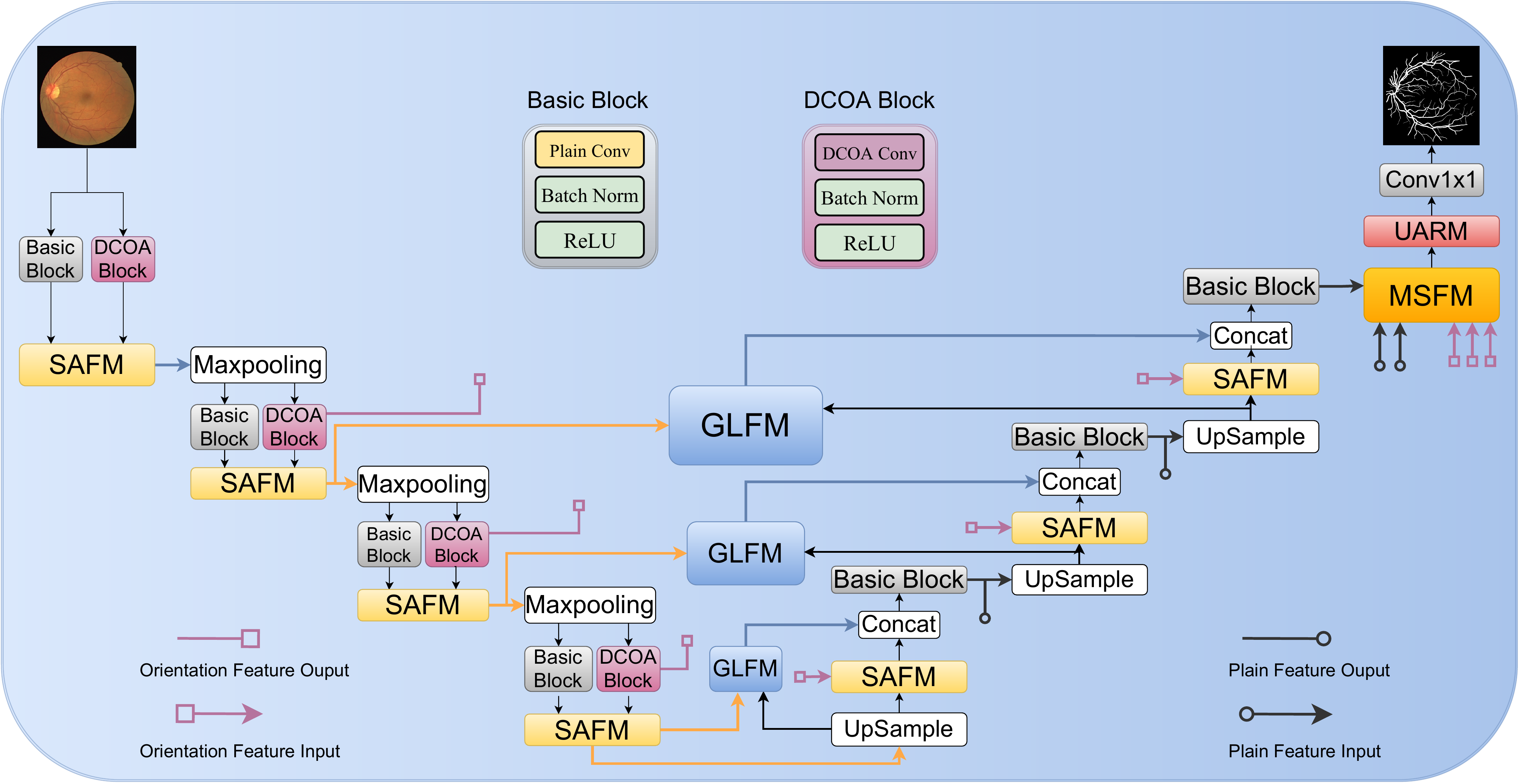}
	\caption{The network architecture of the proposed Orientation and Context Entangled Network (OCE-Net). Plain Conv and DCOA Conv are used to extract the plain and the orientation-aware features, and the SAFM is used for fusing these two features, both of which are essential for extracting the information of vessels. GLFM and MSFM are used to capture context information and entangle the plain and orientation together. UARM is used to focus more attention on the thin vessels.}
	\label{model}
\end{figure*}

The remaining parts of this paper are organized as follows. In Section \ref{sec:related works}, we describe some related works on vessel segmentation, self-attention mechanism and convolutional operators. In Section \ref{sec:methods}, we present a novel framework named OCE-Net and its modules. In Section \ref{sec:data}, we introduce the retinal fundus datasets and the evaluation metrics used in this work.
In Section \ref{sec:experiment}, we conduct extensive ablation and comparison experiments and analyze the function of each module in detail. In Section \ref{sec:conclusion}, a comprehensive conclusion is presented and the future work is discussed.

\section{Related Works}
\label{sec:related works}

\subsection{Traditional Vessel Segmentation Methods}
Many traditional methods have been proposed for vessel segmentation. Among these methods, some of them tried to leverage orientation aware filters to capture the orientation features of vessels.
For example, Zhang et al. \cite{zhang2010retinal} detected retinal blood vessels by matched filter with the first-order derivative of Gaussian (FDOG), which leveraged the orientation selectivity of matched filter.
Soares et al. \cite{soares2006retinal} adopted a multi-scale two-dimensional (2D) Gabor wavelet transform to extract features and used Bayesian classifier to classify each pixel as vessel or non-vessel, which is the first work to introduce Gabor filter into vessel detection.
Azzopardi et al. \cite{azzopardi2015trainable} proposed trainable COSFIRE filters to achieve orientation selectivity.
All these works took the orientation into consideration, however, none of them can handle more than one orientation at a time.
Inspired by these works, we aim to develop a module that can capture complex orientations of vessels.

There are other traiditional methods, for example, 
HED \cite{xie2015holistically} proposed an effective edge detection algorithm which can be introduced to conduct vessel segmentation. 
By taking vessel segmentation as line detection task, LineDet \cite{ricci2007retinal} proposed a vessel segmentation method by using line operators and support
vector machine (SVM) for classification.
Based on LineDet, MS-LineDet \cite{nguyen2013effective} applied line detectors at varying scales and changed the length of the basic line operators. These line detection based methods have the advantage of improving vessel connectivity, but they always misdetect other tissues as blood vessels easily.

In general, these traditional methods leverage the intrinsic characteristics of vessels and achieve good performance. However, the features extracted by these traditional methods lack of discriminability, so they always fail to distinguish capillaries from the fundus background, which results in failure on the detection of thin vessels.

\subsection{Deep Learning Vessel Segmentation Methods}
Deep learning methods for retinal vessel segmentation outperform traditional methods when successfully trained on large-scale datasets with manual labels. 
Among them, there are many groundbreaking works, for example,
DRIU \cite{maninis2016deep} used a basic network and designed two specialized layers to perform blood vessel and optic disc segmentation, which is a pioneering work to introduce deep learning into retinal vessel segmentation.
DeepVessel \cite{fu2016deepvessel} viewed the vessel segmentation as a boundary detection task and introduced Conditional Random Field (CRF) to capture long-range interactions between pixels, which is the first attempt to take context information into consideration.
V-GAN \cite{son2017retinal} introduced Generative Adversarial Networks (GANs) into vessel segmentation to extract clear and sharp vessels with less false positives, which is the first work to introduce GAN into retinal vessel segmentation.
VGN \cite{shin2019deep} incorporated graph convolutional network into CNN to learn the graphical connectivity of vessels.

In addition, some works tried to improve the plain UNet by introducing some novel modules. For example,
Attention UNet \cite{oktay2018attention} introduced an attention gate into UNet for better feature learning and integration with channel attention, in contrast, SA-UNet \cite{guo2021sa} tried to introduce spatial attention into UNet.
DUNet \cite{jin2019dunet} introduced deformable convolution into UNet to adaptively fit the shape of vessels, however, the deformable convolution is very computation-intensive. Inspried by DUNet, a module (DCOA Conv) that can fit the orientation of vessels will be developed in this work.

Other works were dedicated to segment thick and thin vessel seperately.
JL-UNet \cite{yan2018joint} proposed a segment-level loss to focus more on the thickness consistency of thin vessels. 
Yan et al. \cite{yan2018three} proposed a three-stage model to segment thick and thin vessels separately for addressing the imbalance problem between them. 
These works inspried us to develop a module (UARM) that can focus attention seperately on thick and thin vessels.

Moreover, there are some works that aimed at capturing multi-scale features of vessels. For example,
in order to deal with the varying widths and directions of vessel structures, SWT-FCN \cite{oliveira2018retinal} proposed a multiscale Fully Convolutional Neural Network by combining the multiscale analysis and using the Stationary Wavelet Transform.
BTS-DSN \cite{guo2019bts} proposed a multi-scale deeply supervised network with short connections for vessel segmentation. 
CTF-Net \cite{wang2020ctf} proposed a coarse-to-fine SegNet for preserving multi-scale features information.
CC-Net \cite{feng2020ccnet} proposed a cross-connected convolutional neural network to better learn features. These works inspried us to propose the OCE-DNL that can leverage the multi-scale features by fusing them together.

In addition, some methods tried to improve the connectivity of vessels. For example,
DeepDyn \cite{khanal2020dynamic} proposed a stochastic training scheme for deep neural networks in order to balance the precision and recall.
To improve the continuity of thin vessels, SkelCon \cite{tan2022retinal} proposed a light-weight network by introducing skeletal prior and contrastive loss during the training.

Even retinal vessel segmentation performance are largely improved, all of the above methods ignored the complex orientations of vessels and the context information of the whole fundus images, which are improtant for maintaining the continuity and connectivity of vessels.

\subsection{Non-local and Self-Attention Mechanism}
Non-local (NL) \cite{wang2018non} and Self-Attention (SA) \cite{vaswani2017attention} were proposed to model the long-range dependencies without the distance constraints of pixels. Both of them calculated the affinities between the key and query vectors obtained from the same features for capturing the self-correlated attention map. Non-local neural network \cite{wang2018non} computed the response at a position by calculating the weighted sum of the features at all positions in images inspired by NL applied in image denoising \cite{buades2005non}. 

There are some works aiming to improve the performance of NL. For example,
GCNet \cite{cao2019gcnet} found the relationship of Non-local and SE block \cite{hu2018squeeze} and combined them to further improve NL. Disentagled non-local (DNL) \cite{yin2020disentangled} used whitening operator to disentangle the vanilla NL into the pairwise and unary branches for better learning the within-region clues and the salient boundaries. Both of them got insight of NL and improved NL successfully.
Self-attention (SA) mechanism \cite{vaswani2017attention} was proposed in Transformer \cite{vaswani2017attention} to capture the global context information of sequence embeddings in natural language processing (NLP). 

However, Non-local is very computationally expensive. In order to reduce the computationl cost of NL, Ccnet \cite{huang2019ccnet} aimed to reduce the computation complexity of self-attention. 
In addition, in order to further extend the ability of SA, Bello et al. \cite{bello2019attention} combined the vanilla convolution and self-attention to obtain better performance. 
Furthermore, Ramachandran et al. \cite{ramachandran2019stand} presented that self-attention can serve as an effective stand-alone layer, which was a very interesting discovery.
DAN \cite{fu2019dual} introduced self-attention into scene segmentation for adaptively integrating local features with their global dependencies considering the local and global features are equally important, whose motivation is in line with ours. 

In this paper, we adopted Non-local and Self-Attention for capturing context information of fundus images and extended the capacity of non-local by introducing cross-correlation mechanism into the vanilla non-local.

\subsection{Variants of Convolution Operator}
Convolution is a basic operator in CNN for extracting deep representative features \cite{szegedy2015going} \cite{simonyan2014very}. Based on vanilla convolutions, a number of efficient variants of convolution operator have been designed to extend the representation ability of the vanilla convolution.

For instance, some methods aimed to enable the plain convolution operators with the ability of learning the shape of objects. For example,
Active Convolution \cite{jeon2017active} adaptively learned the shape of convolution during training.
Deformable Convolution \cite{dai2017deformable} \cite{zhu2019deformable} learned the offsets of the kernel shapes to fit the object shapes in ROI. Dynamic Region-Aware Convolution \cite{chen2021dynamic} learned to apply each convolution kernel on a single patch region to handle the complex and variable spatial distribution.

Others tried to endow the plain convolution operators with ability of learning the orientation information of objects, for example,
Gabor Convolution \cite{luan2018gabor} introduced Gabor filters with different orientations into vanilla convolution, but it can only learn a single orientation per channel and cannot capture complex orientations. 

In addition, some methods dedicated to endow the plain convolution operators with the ability of integrating several kernels. For example,
CondConv was firstly proposed to integrate several convolution kernels and it \cite{yang2019condconv} learned specialized convolutional kernels for each example by introducing conditional parameters. Dynamic Convolution was further improved by introducing attention mechanism to learn the weights and the biases of each vanilla convolution and integrated them into a single convolution kernel \cite{chen2020dynamic}. 

All of the above convolution operators extended the capacities of vanilla convolution, however, most of them do not take the orientation of objects into consideration, and some of them \cite{luan2018gabor} can only encode simple orientation information. In this paper, we proposed a novel convolution which has the ability of capturing complex orientations in order to fully extract the features of fundus vessels for improving the continuity of thin vessels.

\section{Methodology}
\label{sec:methods}

\subsection{Overall Architecture}

The network architecture of the proposed OCE-Net is shown in Fig. \ref{model}. 
The backbone of OCE-Net is the vanilla UNet \cite{ronneberger2015u}, which is widely used in medical image segmentation. In the down-sampling stage, we use the proposed Dynamic Complex Orientation Aware Convolution (DCOA Conv) to extract the complex orientation features of vessels and employ the plain convolution to extract the plain features of the fundus images, and then a Selective Attention Fusion Module (SAFM) is utilized to fuse these two kinds of features (the reason for which will be explained in Section. \ref{conflict}). A Global and Local Fusion Module (GLFM) is proposed to play the role of attention gate of UNet for simultaneously capturing the global and local features of vessels. In the up-sampling stage, the extracted orientation aware features are used as the prior guidance for improving the continuity. At the end of the network, a Multi-Scale Fusion Module (MSFM) is introduced with an Orientation and Context Entangled DNL (OCE-DNL) to entangle the orientation and context information together. Finally, an Unbalanced Attention Refining Module (UARM) is proposed to refine the output feature by focusing more attention on thin vessels.


\subsection{Dynamic Complex Orientation Aware Convolution}
\begin{figure}[htbp]
	\centering
	\includegraphics[width=9cm]{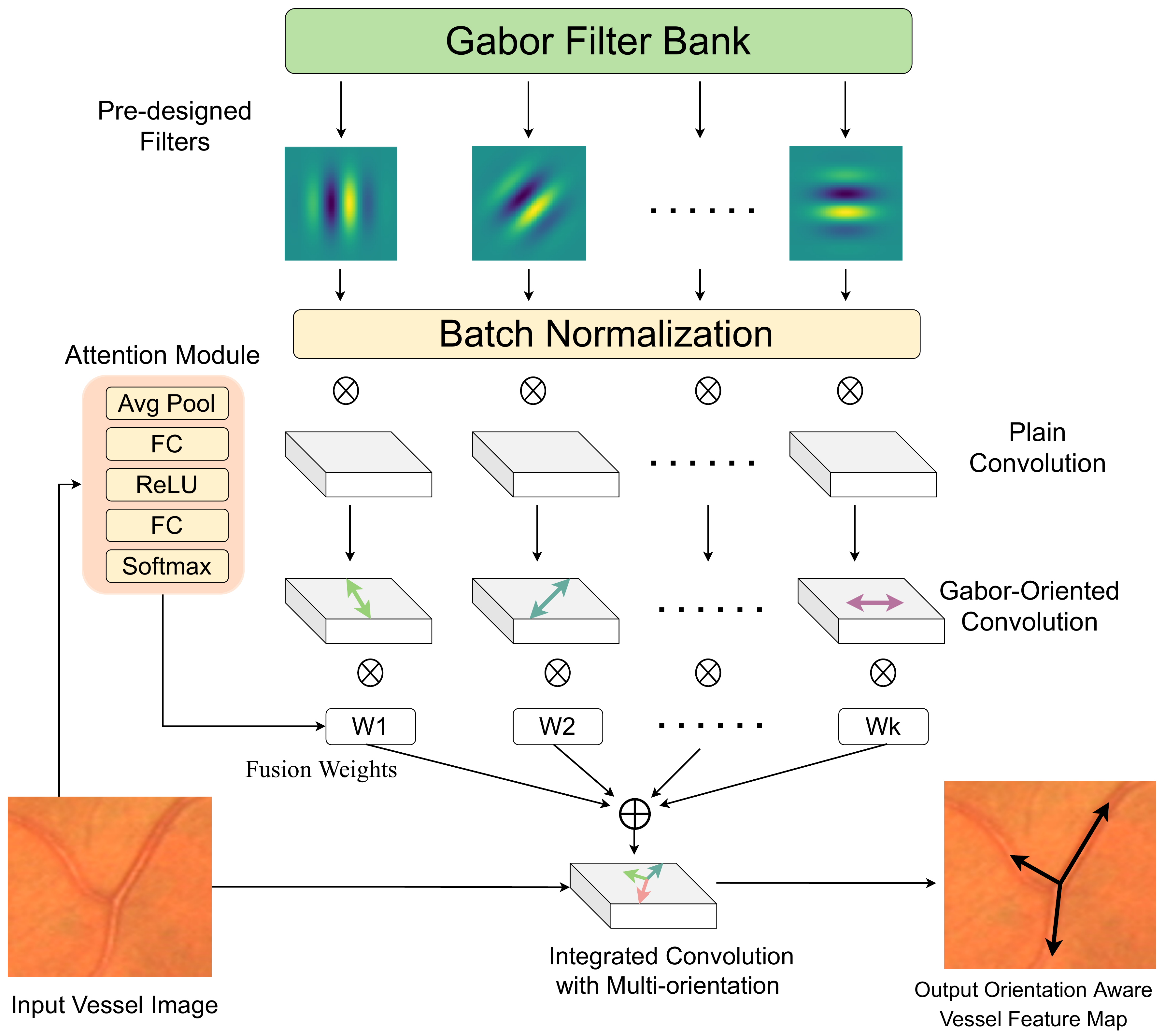}
	\caption{The proposed Dynamic Complex Orientation Aware Convolution (DCOA Conv).}
	\label{dcoa}
\end{figure}

Orientation of tissues is an important feature for medical image segmentaiton \cite{soares2006retinal} \cite{cherukuri2019deep}. In order to capture the complex orientation information of the blood vessels, a Dynamic Complex Orientation Aware Convolution (DCOA Conv) is proposed to extract the features of vessels with multiple orientations within the same receptive field. 
As shown in Fig. \ref{dcoa}, in the proposed DCOA Conv, the oriented Gabor filters are generated from a pre-designed Gabor Filter Bank. The 2D Gabor function is defined as \cite{yuan2022adaptive}

\begin{equation}
	G(x,y;\lambda , \theta , \psi , \sigma , \gamma ) = exp(-\frac{x^{'2}+\lambda ^{2}y^{'2}}{2\sigma ^{2}})exp(i(2\pi \frac{x^{'}}{\lambda})+\psi)
\end{equation}

where $x^{'} = xcos(\theta) + ysin(\theta)$ and $y^{'} = - xsin(\theta) + ycos(\theta)$. $x$ and $y$ are respectively the horizontal and vertical coordinates of pixels.  $\lambda$ indicates the wavelength of the gabor filter and it was set to 1/$\sqrt{2}$. $\theta$ represents the orientation of a filter. $\psi$ is the phase offset and it was set to 0. $\sigma$ denotes the standard deviation and it was set to 1. $\gamma$ is the spatial aspect ratio and it was set to 1. As shown in Fig. \ref{dcoa}, by setting different values for $\theta$, filters with different orientations can be obtained. There are filters with 8 orientations, so the values of $theta$ were set to {2*$\pi$ / i, (i=1,2,...7,8)}.
Then these kernels are multiplied with eight vanilla convolution kernels, which aims to assign the filter kernels with orientation preference to the vanilla convolution whose kernel has no orientation selectivity. A batch normalization layer \cite{ioffe2015batch} is used here for normalizing the weights of the Gabor kernels. Following the work in \cite{chen2020dynamic}, an attention module is used to learn the weight coefficients of each oriented kernel for selecting the useful convolution operators and the attention coefficients were computed following the design in \cite{chen2020dynamic}. Note that the selection depends on the orientation of the vessels in the receptive field. If there is no vessel along the orientation of the receptive field, the weight coefficient of this oriented convolution would be set to 0 by the attention module and it would not be integrated into the final convolution kernel with multiple orientations.

Then the selected convolution kernels are integrated together to form a single convolution operator which has a kernel with multiple selective orientations. The DCOA Conv is written as

\begin{equation}
	DCOA = \sum_{i=1}^{8}w_{i}(K_{i}\otimes BN(G(\theta_{i}))) \quad (i=1,2,3,...8)
\end{equation}

where $G(\theta_{i})$ is the composite Gabor kernels. $BN$ denotes the Batch Normalization layer. $K_{i}$ is the plain convolution kernel without orientation. $\otimes$ means multiplication operator. $w_{i}$ denotes the weight coefficient learned by the attention module. Our experiments have shown that choosing eight ($i=8$) orientations can encode the orientations of all blood vessels well. 
This composite convolution kernel has the ability of capturing the vessels with complex multiple orientations in a single receptive field. 
It is dynamic because the orientations of the final composite convolution kernel can be dynamically adjusted by learning to set different weight coefficients for different oriented convolution kernels based on the orientations of specific vessels.

As shown in Fig. \ref{model}, a DCOA Block now can be obtained by stacking DCOA Conv, Batch Normalization (BN) and ReLU function together for extracting orientation features $F_{o}$:
\begin{equation}
	\textit{F}_{o} = ReLU(BN(DCOA(F_{in})))
\end{equation}
where $F_{in}$ means the input features.
From Fig. \ref{model}, the basic block can be obtained by directly replacing the DCOA Conv in DCOA block with the vanilla convolution to extract the plain features without orientation preference. The kernel sizes of both vanilla convolution and DCOA Conv are set to 3x3 in their own blocks.

The DCOA Conv in DCOA block is used to capture complex orientations of vessels, however, 
DCOA Conv only focuses on the vessels along one or several orientations and ignores other important features that are not located at these particular orientations, such as the fundus background and other non-vascular tissues, which are not oriented but also helpful for identifying blood vessels when serving as the important negtive samples. Therefore, the best solution is to capture both the plain features and the orientation features extracted by the basic blocks and the DCOA blocks described above. In other words, the extracted orientation features should be viewed as the auxiliary prior, which are used to guide the segmentation of blood vessels.

\begin{figure}[htbp]
	\centering
	\includegraphics[width=10cm]{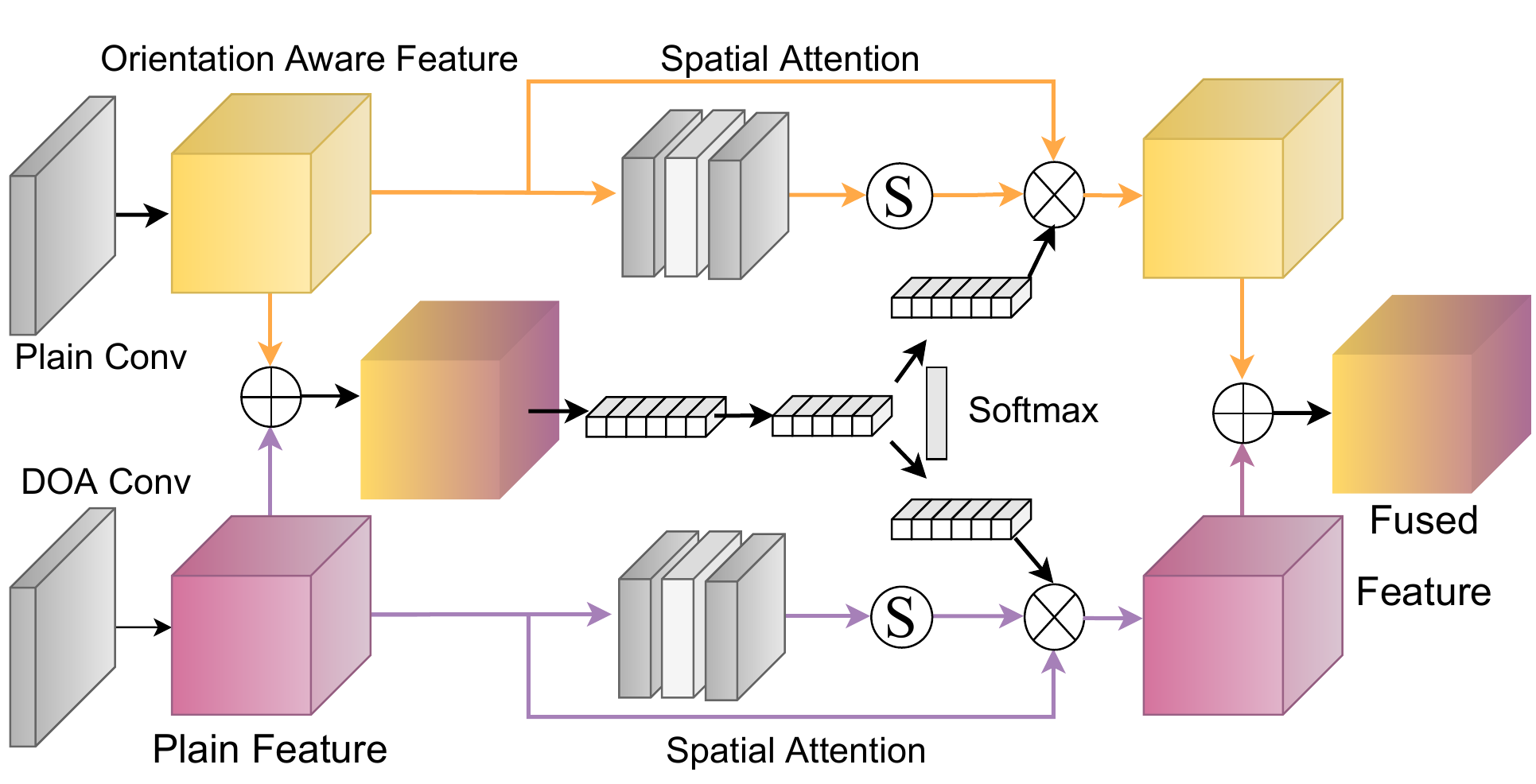}
	\caption{The schematic of the Selective Attention Fusion Module (SAFM). The plain features and orientation features are fused with the help of channel-wise selection and spatial-wise attention. This SAFM is inspired by the the framework of selective kernel network 
		(SKNet) in \cite{li2019selective}.}
	\label{safm}
\end{figure}

As shown in Fig. \ref{safm}, inspired by \cite{li2019selective}, a Selective Attention Fusion Module (SAFM) is introduced to fuse the plain and orientation features, which are respectively extracted by the basic block and the proposed DCOA block. Channel attention mechanism is used to select useful channels of both two kinds of features before fusion, because there are some redundant channels in both plain and orientation features of fundus images, which contain much noise and artifacts. The spatial attention \cite{qin2020ffa} is used to focus more attention on the useful features in both two branches, because some tissues like exudates, hemorrhages and maculas have similar features with blood vessels, which may mislead the network to identify them as blood vessels.

The spatial attention $SPA$ is defined as
\begin{equation}
	F_{spa} = SPA(F_{in}) = F_{in} * \delta(Conv(ReLU(Conv(F_{in}))))
\end{equation}
where $F_{in}$ and $F_{spa}$ are respectively the input and output features of spatial attention. $Conv$ means the convolution operator a kernel of 3x3 size. $ReLU$ means the ReLU function. $\delta$ denotes the sigmoid function.

\begin{figure}[htbp]
	\centering
	\includegraphics[width=13cm]{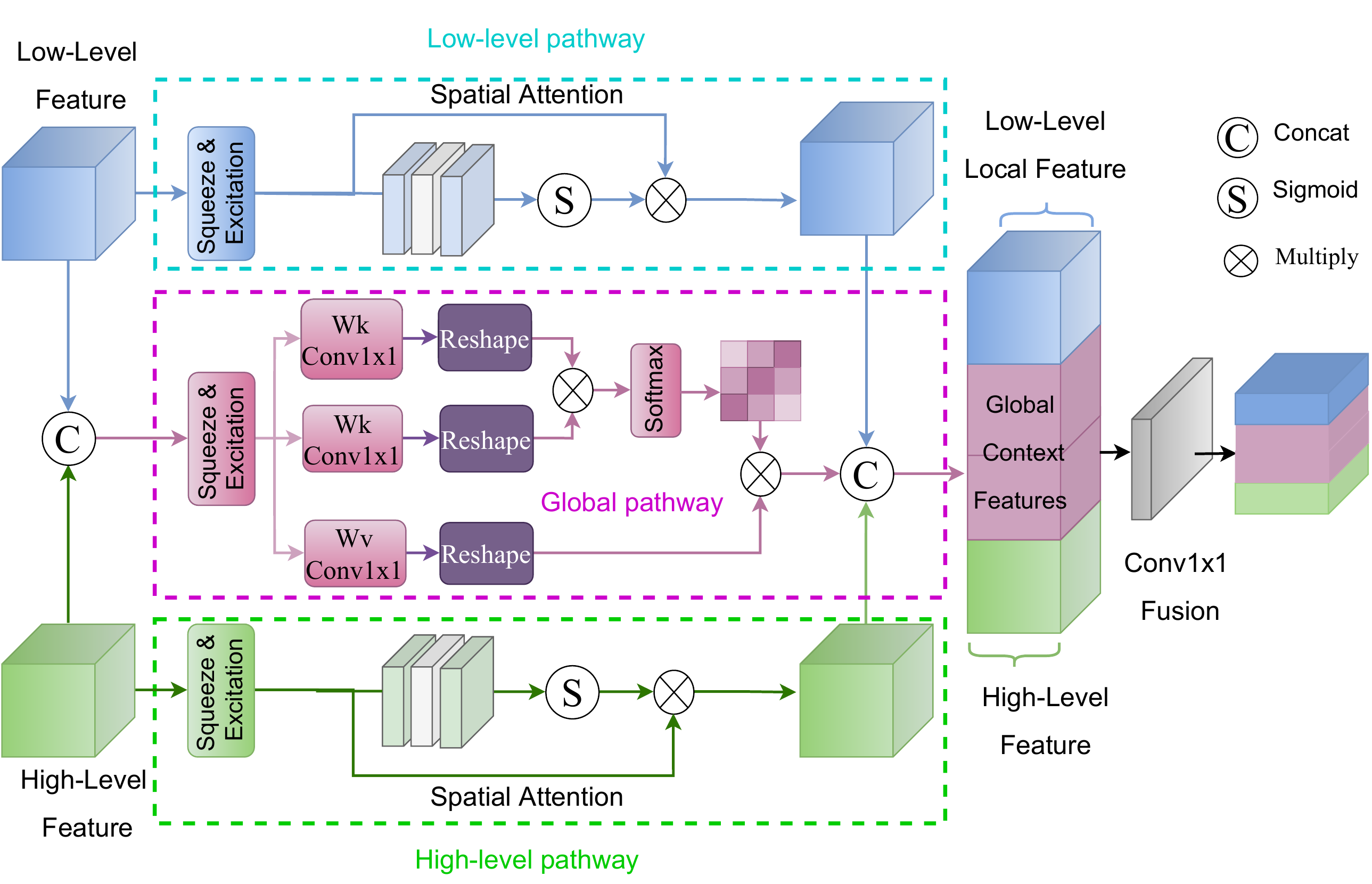}
	\caption{The proposed Global and Local Fusion Module (GLFM).}
	\label{glfm}
\end{figure}

\subsection{Global and Local Fusion Module}
Convolution with the kernel of 3x3 size has a local receptive field and has been proved very important for many computer vision tasks.
But such convolution can not capture global contextual information of the whole input image, which is also very helpful for image segmentation \cite{wang2018non} \cite{fu2019dual}. 
Retinal vessels have abundant context information \cite{wang2020csu}, for example, the complex furcations and branches of blood vessels have their relative positions and sizes against the entire vascular system, which present a kind of symmetry from the view of the whole retinal fundus image. Furthermore, some local lesion areas may occlude the blood vessels, and the global context information can help reconstruct the occluded vessels and improve the continuity of vessels from a holistic view.
In addition, thin vessels contain rich local detail information and has a more complex orientation diversity. In a word, capturing global context and local detail information are equally important for vessel segmentation. A Global and Local Fusion Module (GLFM) is proposed herein to achieve this goal.

As shown in Fig. \ref{glfm}, GLFM is composed of three pathways, i.e., low-level, high-level and global pathways.
The squeeze-and-excitation operations $SE(.)$ \cite{hu2018squeeze} are used here for channel-wise integration because there is noise and other interference in the retinal fundus image. In both the low-level and high-level pathways, spatial attention $SPA(.)$ \cite{qin2020ffa} with a 3x3 local convolution is used for focusing more attention on some local areas and vessel tissues. In addition, in the global pathway, both the low-level $F_{L}$ and high-level $F_{H}$ features are concatenated via $Concat(.)$ and used to model the global long-range dependencies via self-attention $SA(.)$ mechanism, then the global context features $F_{G}$ are obtained. 
\begin{equation}
	\begin{aligned}
		&F_{LL} = SPA(SE(F_{L})) 
		\\&F_{HH} = SPA(SE(F_{H})) 
		\\&F_{G} = SA(SE(Concat(F_{H}, F_{L})))
	\end{aligned}
\end{equation}
where $F_{LL}$, $F_{HH}$ and $F_{G}$ denote the low-level local features, high-level features and global context features, respectively.

Finally, these three kinds of features are concatenated via $Concat(.)$ and fused via a 1x1 convolution $Conv(.)$ to contain the output features of $F_{glfm}$ which simultaneously captures the global context information and the local detail information.
\begin{equation}
	F_{glfm} = Conv(Concat(F_{LL}, F_{HL}, F_{G}))
\end{equation}

\begin{figure*}[htbp]
	\centering
	\includegraphics[width=14cm]{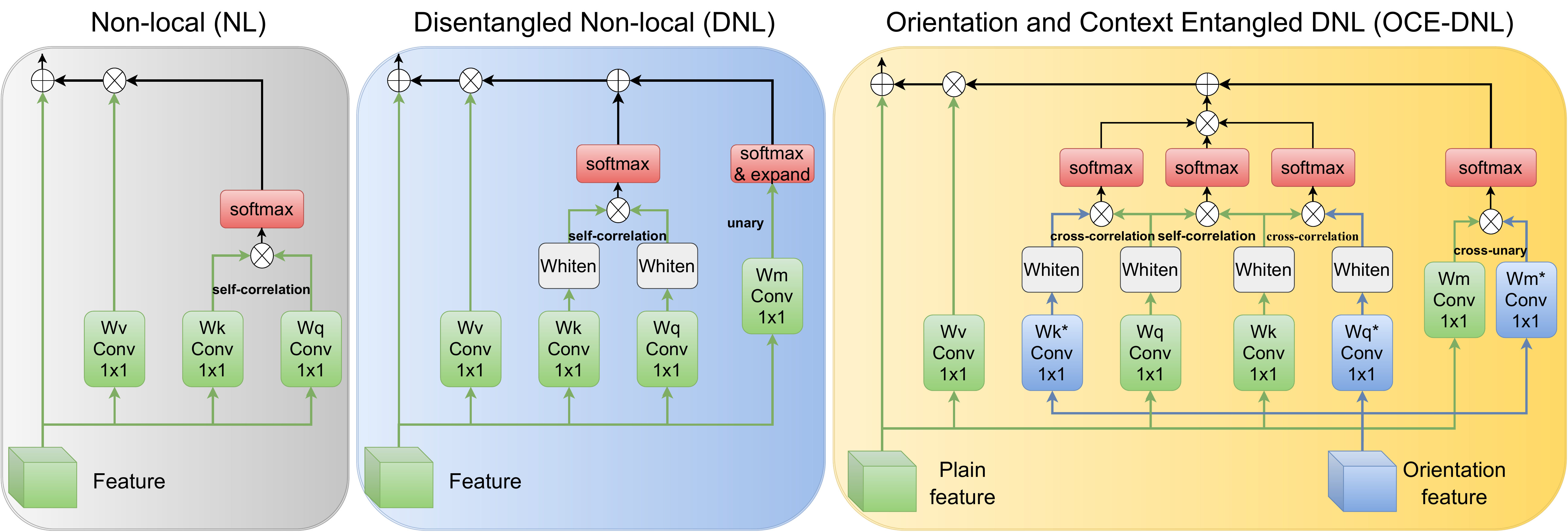}
	\caption{The schematic of Non-local (NL) \cite{wang2018non}, Disentangled Non-local (DNL) \cite{yin2020disentangled} and the proposed Orientation and Context Entangled Non-local (OCE-DNL). Due to space limitation, OCE-NL is not presented here. 'Whiten' here means whitening operator. Note that the functions of whiten operator in DNL are to: 1. reduce the correlation between features. 2. make features have the same variance. More details about the whitening operator can be found in \cite{yin2020disentangled}.}
	\label{ocenl}
\end{figure*}

\subsection{Orientation and Context Entangled NL/DNL}
In order to fully mine the association between plain and orientation features, inspired by Disentangled non-local (DNL) \cite{yin2020disentangled}, an Orientation and Context Entangled DNL (OCE-DNL) is proposed to entangle the orientation and context information together, in which orientation information is used to guide MSFM to better learn context information and discriminate blood vessels from the plain features.
The NL and DNL are defined as
\begin{equation}
	\begin{aligned}
		&X_{out}^{NL}(x)  = X_{in}(x) + W_{NL}(x_{i}, x_{j})\cdot V(X_{in}(x)) \\ \\
		&X_{out}^{DNL}(x) = X_{in}(x) + W_{DNL}(x_{i}, x_{j})\cdot V(X_{in}(x))
	\end{aligned}
\end{equation}
where $X_{in}(x)$ means the input features and $x$ denotes the position of pixel in the features. $V$ means the value vector of NL. $W_{NL}$ and $W_{DNL}$ represent the self-correlation weights (self-attention map) yielded by computing the affinities between the Query and Key vectors, respectively, which are defined as
\begin{equation}
	\begin{aligned}
		&W_{NL}(x_{i}, x_{j}) = Q_{i}^{T}\cdot K_{j}  \\ \\
		W_{DNL}(x_{i}, x_{j}) &= (Q_{i} - \mu_{Q})^{T}\cdot (K_{j} - \mu_{K}) + \mu_{Q}\cdot k_{j}
	\end{aligned}
\end{equation}
where $Q_{i}$ and $K_{j}$ denote respectively the query and key vectors in NL. $(.)^{T}$ means the transposition operator. $\mu_{Q}$ and $\mu_{K}$ denote their mean values calculated by a whitening operation in DNL. $x_{i}$ and $x_{j}$ represent two different pixels in the input features.

We propose a novel non-local operator to explore the potential association between two related features by calculating their cross-correlation. To our best knowledge, this is the first attempt to introduce cross-correlation (cross-attention) into the non-local that was originally designed to capture only self-correlation or self-attention.
The proposed cross-correlation based entanglement is defined as
\begin{equation}
	\begin{aligned}
		W_{OCE-NL}(x_{i}, x_{j};y_{i}, y_{j}) 
		= (Q_{i}^{T}\cdot K_{j})
		\cdot {\color{blue}(\tilde{Q_{i}}^{T}\cdot K_{j})
			\cdot (Q_{i}^{T}\cdot \tilde{K_{j}})} \\
	\end{aligned}
	\label{eq_nl}
\end{equation}
\begin{equation}
	\begin{aligned}
		\\&W_{OCE-DNL}(x_{i}, x_{j}; y_{i}, y_{j}) 
		= [(Q_{i} - \mu_{Q})^{T}\cdot (K_{j} - \mu_{K})] \\
		&\cdot {\color{blue}[(\tilde{Q_{i}} - \tilde{\mu_{Q}})^{T}\cdot (K_{j} - \mu_{K})]}
		\cdot {\color{blue}[(Q_{i} - \mu_{Q})^{T}\cdot (\tilde{K_{j}} - \tilde{\mu_{K}})]}   \\
		&+ (\mu_{Q}\cdot k_{j}) \cdot {\color{blue}(\tilde{\mu_{Q}}\cdot \tilde{k_{j}})}
	\end{aligned}
	\label{eq_dnl}
\end{equation}

where $Q_{i}$ and $K_{i}$ denote respectively the query and key vectors of the plain features. $\tilde{Q_{i}}$ and $\tilde{K_{i}}$ denote respectively the query and key vectors of the orientation features. $\mu_{Q}$ and $\mu_{K}$ denote respectively the mean values of plain features which are calculated by a whitening operation \cite{yin2020disentangled}. $\tilde{\mu_{Q}}$ and $\tilde{\mu_{K}}$ denote the mean values of orientation features calculated by a whitening operation. $x_{i}$ and $x_{j}$ represent two different pixels in the plain features and $y_{i}$ and $y_{j}$ represent two different pixels in the orientation features.
Note that the blue parts of Eq. \ref{eq_nl} and \ref{eq_dnl} represent cross-correlation calculations.

As shown in Fig. \ref{ocenl}, the multi-scale plain features outputted by the network and the multi-scale orientation features extracted in the down-sampling stage are decoupled into query and key vectors by 1x1 convolutions. As for the original non-local \cite{wang2018non}, only the self-correlation between the query and key of the plain feature is computed to obtain a self-attention map by multiplying them, and this self-attention map can model long-range dependencies and capture global context information. In order to calculate the cross-correlation between two different features, the respective query and key vectors for the two features are multiplied separately and the cross-attention maps can be obtained. Then the self-attention map is used to capture the context information in the plain features and the cross-attention map is used to model the global relationship between the plain features and the prior orientation features. Through computing cross-attention and applying cross-attention maps to the plain features, the context and orientation information can be entangled together into the final output features, which are ultimately used to predict the vessels from the fundus background.

\begin{figure}[htbp]
	\centering
	\includegraphics[width=11cm]{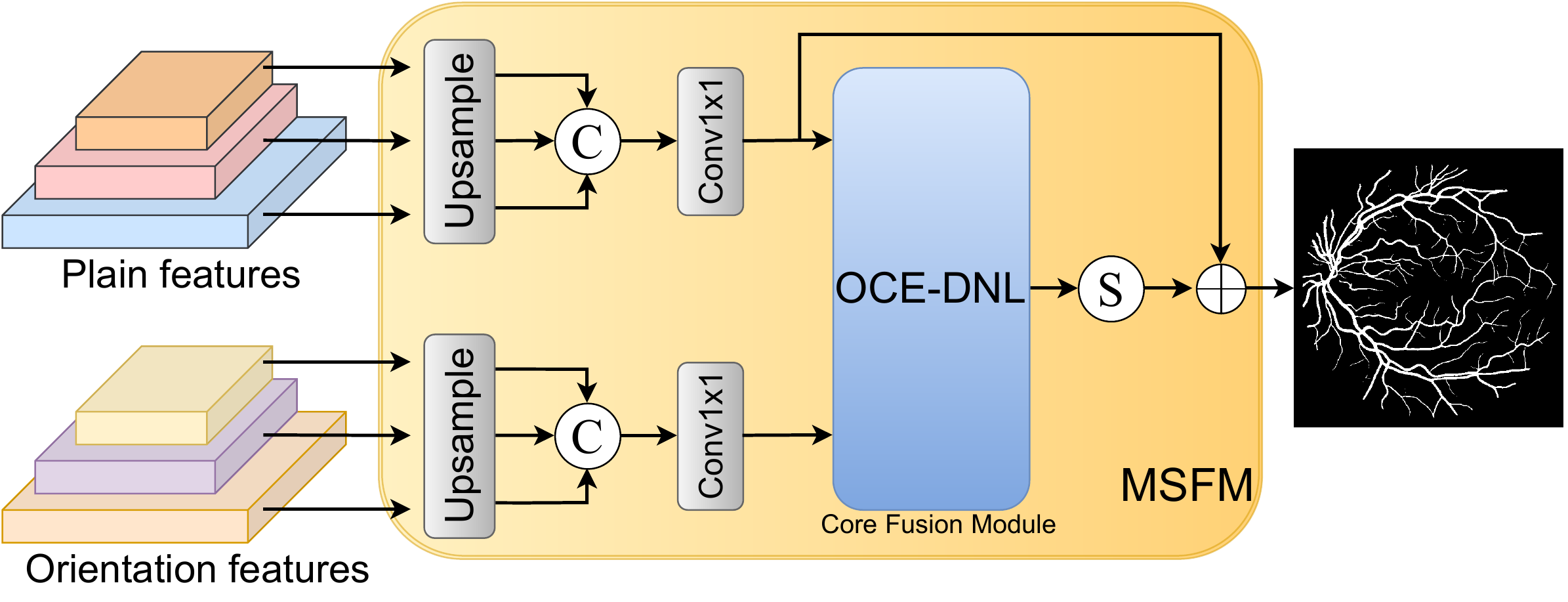}
	\caption{The proposed Multi-Scale Fusion Module (MSFM).}
	\label{msfm}
\end{figure}

Leveraging multi-scale features is helpful for better vessel segmentation. Inspired by \cite{wu2021region}, we design a Multi-Scale Fusion Module (MSFM) to fuse multi-scale features including multi-scale plain features $F_{i} \in \mathbb{R}^{\frac{C}{r} \times H \times W}, (i,r=1,2,3)$ and multi-scale orientation features $F_{i}^{o} \in \mathbb{R}^{\frac{C}{r} \times H \times W}, (i,r=1,2,3)$ , as shown in Fig. \ref{msfm}. The features at different scales are unified through up-sampling $Up_{i}(.) (i=1,2,3)$ and concatenated in the channel dimension. Then these features are fused and dimensionally reduced through an 1x1 convolution, yielding the plain input feature $F_{in}$ and the orientation input features $F_{in}^{o}$ for MSFM as follows 
\begin{equation}
	\begin{aligned}
		F_{in} = Conv(Concat(Up_{1}(F_{1}), Up_{2}(F_{2}), Up_{3}(F_{3})))  	\\ \\
		F_{in}^{o} = Conv(Concat(Up_{1}(F_{1}^{o}), Up_{2}(F_{2}^{o}), Up_{3}(F_{3}^{o})))
	\end{aligned}
\end{equation}
The core fusion module of MSFM is the proposed OCE-DNL, which is used to entangle the orientation and context information together by computing cross-correlation between the plain and the prior orientation features. And the output feature $F_{msfm}$ after fusion can be obtained by entangling the context and orientation information as follows
\begin{equation}
	F_{msfm} = \delta (OCEDNL(F_{in}, F_{in}^{o})) + F_{in}
\end{equation}
where $\delta$ means the sigmoid function and $OCEDNL$ denotes the proposed OCE-DNL module.

\subsection{Unbalanced Attention Refining Module}
There are two kinds of unbalance in fundus images. One is the serious unbalance between the pixel numbers of blood vessels and the fundus background. The fundus background occupies the majority of pixels, while the blood vessels only take up a small proportion of total pixels. The other is the unbalance between the numbers of thick and thin vessels \cite{yan2018three}. Thick vessels are generally large in width and hence occupy the majority of the blood vessels, while the width of thin vessels is usually 3-5 pixels. These unbalances make blood vessels difficult to be detected and identified from the background, and also make thin vessels harder to be detected due to their less prominent features. In order to deal with these unbalances, previous deep learning methods \cite{khanal2020dynamic} usually design class-balanced loss or introduce weighted coefficients into the pixel-wise loss functions for imposing more penalties on thin vessels \cite{yan2018joint}. Here we propose a novel approach to tackle this these unbalances from the perspective of visual attention mechanism.

\begin{figure}[htbp]
	\centering
	\includegraphics[width=12cm]{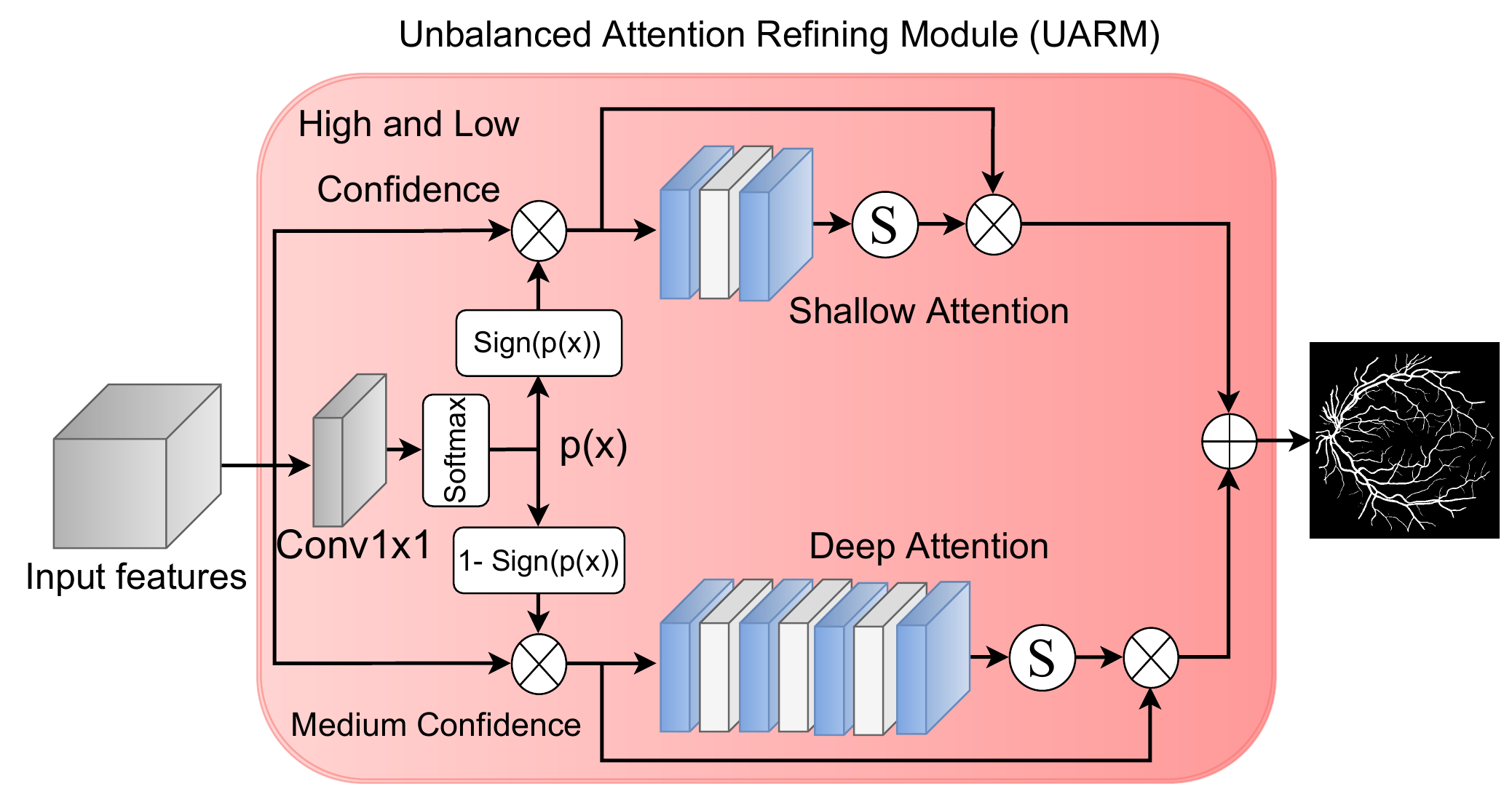}
	\caption{The proposed Unbalanced Attention Refining Module (UARM). The blue blocks represent the Conv3x3 and the gray blocks denote the ReLU function. The deeper attention module is adopted for mining hard sample. Note that the output channel number of the last convolutional block in both deep and shallow paths is 1 for calculating the probability map.}
	\label{uarm}
\end{figure}

As shown in Fig. \ref{uarm}, a novel Unbalanced Attention Refining Module (UARM) works to focus more visual attention on the vessels, especially the thin vessels. The proposed UARM is applied at the end of the network to refine the final output features of the network. Instead of directly applying spatial attention to the final output feature (the input feature $F_{in} \in  \mathbb{R}^{C\times H\times W}$ of UARM), we first use a 1x1 convolution to reduce the dimension of the feature from 32 to 1 and use the softmax function to obtain a probability map $p(F_{in})$, which describes the probability that each pixel belongs to a vessel, ranging from 0 to 1. 
\begin{equation}
	p(F_{in}) = \textit{Softmax}(Conv(F_{in}))
\end{equation}
Then, a pre-defined Sign function $Sign(.)$ is used to separate the probability values into three intervals, corresponding to three different regions in the image. In other words, we divide the image into three different regions of high confidence, medium confidence and low confidence according to the probability of pixels belonging to vessels.
The $Sign(.)$ function is defined as
\begin{equation}
	Sign(x)=
	\begin{cases}
		1 \quad \quad \quad {0 \leq x < 0.4}\\
		0 \quad \quad \quad {0.4 \leq x < 0.7}\\
		1 \quad \quad \quad {0.7 \leq x < 1.0}
	\end{cases}
\end{equation}
where $x$ means the probability of each pixel in the probability map $p(F_{in})$. 

By setting two thresholds between 0 and 1 (these two thresholds are experimently set to 0.4 and 0.7), we can separate the features $F_{2}$ with medium confidence regions. And high and low confidence regions are combined together and separated into features $F_{1}$ as well. $F_{1}$ and $F_{2}$ are fomulated as
\begin{equation}
	F_{1} = F_{in} \otimes Sign(p{F}_{in})),\quad
	F_{2} = F_{in} \otimes (1 - Sign(p(F_{in}))
\end{equation}
We found that the network usually pays more attention to the thick vessels and tends to ignore the thin vessels, that is, less attention is allocated to the ambiguous areas with medium confidence, where thin vessels are always located. Therefore, for such ambiguous regions, a deeper attention module $Att_{d}(.)$ with stronger discrimination ability is solely applied to gain more attention. For the regions with high and low confidence, a shallow attention module $Att_{s}(.)$ is used because thick vessels (usually composed of pixels with high probability) and the fundus background (usually composed of pixels with low probability) are highly discriminable. The process of UARM is defined as
\begin{equation}
	F_{uarm}= \textit{UARM}(F_{in}) = Att_{s}(F_{1}) + Att_{d}(F_{2})
\end{equation}
where $F_{uarm}$ denotes the output features of UARM.

This unbalanced, biased approach for applying attention allows the network to better focus on uncertain areas that need more attention.

\subsection{Loss Function}
A Cross Entropy loss $\mathcal L_{CE}$ is adopted as the loss function of our OCE-Net for vessel segmenation, which is defined as

\begin{equation}
	\mathcal L_{CE}(p,q) =  -\sum_{k=1}^{N}p_{k}\ast log(q_{k})
\end{equation}

\section{Datasets and Materials}
\label{sec:data}

\subsection{Retinal Fundus Datasets}
Our model was trained on three widely-used datasets, including DRIVE \cite{staal2004ridge}, STARE \cite{hoover2000locating} and CHASEDB1 \cite{fraz2012ensemble}. 

The DRIVE dataset contains 40 pairs of retinal fundus images with their corresponding labels, which were manually delineated by two human observers, and the labels of 1st observer are usually used as the ground truth. The size of each fundus image in DRIVE is 565 $\times$ 584 pixels, and the training and test sets are unoverlapped with each other, each of which contains 20 pairs of images. 

The STARE dataset consists of 20 fundus images with their corresponding manual labels annotated by two human experts. The resolution of each image is 700 $\times$ 605 pixels. Generally, the first 10 images and their labels are used as training set, and the rest 10 images are used as test set. 

The CHASEDB1 dataset contains 28 fundus images and their labels with a resolution of 999 $\times$ 960 pixels. The first 20 images are usually used as training set and the rest 8 images are considered as test set.

In order to evaluate the generalization performance of our model, we also tested our OCE-Net on some challenging datasets, including AV-WIDE \cite{estrada2015retinal}, UoA-DR \cite{chalakkal2017comparative}, RFMiD \cite{pachade2021retinal} and UK Biobank \cite{sudlow2015uk}.

The AV-WIDE dataset contains 30 wide-FOV color images and the arteries and veins were annotated separately for artery-vein classification. The resolutions of images vary, but most of them are around 1300 $\times$ 800 pixels. The vessels are usually very thin in AV-WIDE.
The UoA-DR dataset consists of 200 images with a resolution of 2124 $\times$ 2056 pixels, which were collected by University of Auckland. 
The RFMiD dataset contains 3200 fundus images. 1920 images of them are divided into the training set, 640 images are divided into the validation set and the rest 640 images are divided into the test set. The fundus images were captured by three different fundus cameras. The sizes of images vary, having the resolutions of 4288 $\times$ 2848 (277 images), 2048 $\times$ 1536 (150 images) and 2144 $\times$ 1424 (1493 images), respectively.
The UKBB dataset contains 100K fundus images with the size of 2048 $\times$ 1536 pixels.

Note that instead of retraining the model on these datasets, we just tested on these challenging sets using the models already trained on DRIVE \cite{staal2004ridge}.

\subsection{Evaluation Metrics}
We evaluated our model with some frequently-used metrics, including F1 score (F1), accuracy (Acc), sensitivity (SE), specificity (SP) and area under the ROC curve (AUC), which are defined as

\begin{equation}
	\begin{aligned}
		SE = Re&call = \frac{TP}{TP + FN} \qquad SP = \frac{TN}{TN+FP} \\  \\
		&F1 = 2 \times \frac{Precision \times Recall}{Precision + Recall} \\  \\
		&Acc = \frac{TP + TN}{TP + TN + FP + FN} 
	\end{aligned}
\end{equation}
where $TP$, $TN$, $FP$, and $FN$ represent the numbers of true positive,
true negative, false positive, and false negative pixels, respectively. In addition, we also adopted some improved metrics proposed by Gegundez et al. \cite{gegundez2011function}, including connectivity (C), overlapping area (A), consistency of vessel length (L) and the overall metric (F).
The overall metric (F) is defined as
\begin{equation}
	F = C \times A \times L
\end{equation}

Moreover, Yan et al. \cite{yan2017skeletal} proposed some other novel metrics, including rSE, rSP and rAcc, for improving the corresponding SE, SP and Acc metrics, respectively. We also used these newly-designed metrics to evaluate our model and compare with other methods. More details about these redefined metrics can be found in \cite{yan2017skeletal}. Besides that, the Matthews Correlation Coefficient (Mcc) \cite{khan2020hybrid} was also used to evaluate our model.

\section{Experiments}
\label{sec:experiment}

\subsection{Implementation details}
We built our model using PyTorch framework \cite{paszke2019pytorch}. The model was trained on a TITAN XP GPU with 12G memory. The Adam optimizer \cite{kingma2014adam} and the Cosine Annealing Learning Rate (LR) were adopted during the training. Before training, the fundus images were converted from RGB to grayscale, then Gamma correction and CLAHE \cite{pizer1987adaptive} were applied to enhance the lightness and contrast of the grayscale fundus images. The images were randomly cropped into patches with a size of 48 $\times$ 48 pixels due to the limitation of GPU memory and the number of cropped patches was set to 15000. The batch size was set to 32 and the total epoch was set to 50. The early stopping strategy was adopted and the epoch was set to 8.

\subsection{Overall comparison with other methods}
In order to evaluate the performance of our method and demonstrate its superiority, extensive quantitative and qualitative experiments have been conducted.

As shown in Table. \ref{drive}, \ref{stare} and \ref{chasedb}, we compared our method with numerous state-of-the-art methods on the DRIVE \cite{staal2004ridge}, STARE \cite{hoover2000locating} and CHASEDB1 \cite{fraz2012ensemble} datasets. In terms of the commonly used indicators, our method outperforms most of previous state-fo-the-art methods especially on DRIVE. 

Among these compared methods, BTS-DSN \cite{guo2019bts} neither focus more attention on local thin vessels and capture global context information, therefore, it achieved poor performance on both SE and SP. CSU-Net \cite{wang2020csu} developed a context path to capturing the global context of images but neglected to emphasize local details of thin vessel, therefore, it achieved good performance on SP but relatively low in SE. CTF-Net \cite{wang2020ctf} specially designed a Fine segNet to deal with local thin vessels but neglected the global context of the whole vascular system, therefore, it achieved high indicators on SE but low on SP. Different from these methods, our OCE-Net captures both global and local information of vessel as well as focuses more attention on thin vessels, so OCE-Net achieves promising performance on both SE and SP.

As shown in Table. \ref{drive2}, in terms of some newly redefined metrics, our method also outperforms many other recent methods. 

SkelCon \cite{tan2022retinal} adopted the contrastive learning (CL) strategy to better fit the shape of vessel for improving the connectivity of thin vessels, therefore, it achieved better performance on vessel connectivity. However, it did not capture global context information of vessels and focus more attention on thin vessels, so it obtained poor performance on rSE and rSP indicators. In comparison, thanks to taking both local and global information into consideration, our OCE-Net can achieve better performance than SkelCon on both rSE and rSP.

\begin{figure*}[htbp]
	\centering
	\includegraphics[width=14cm]{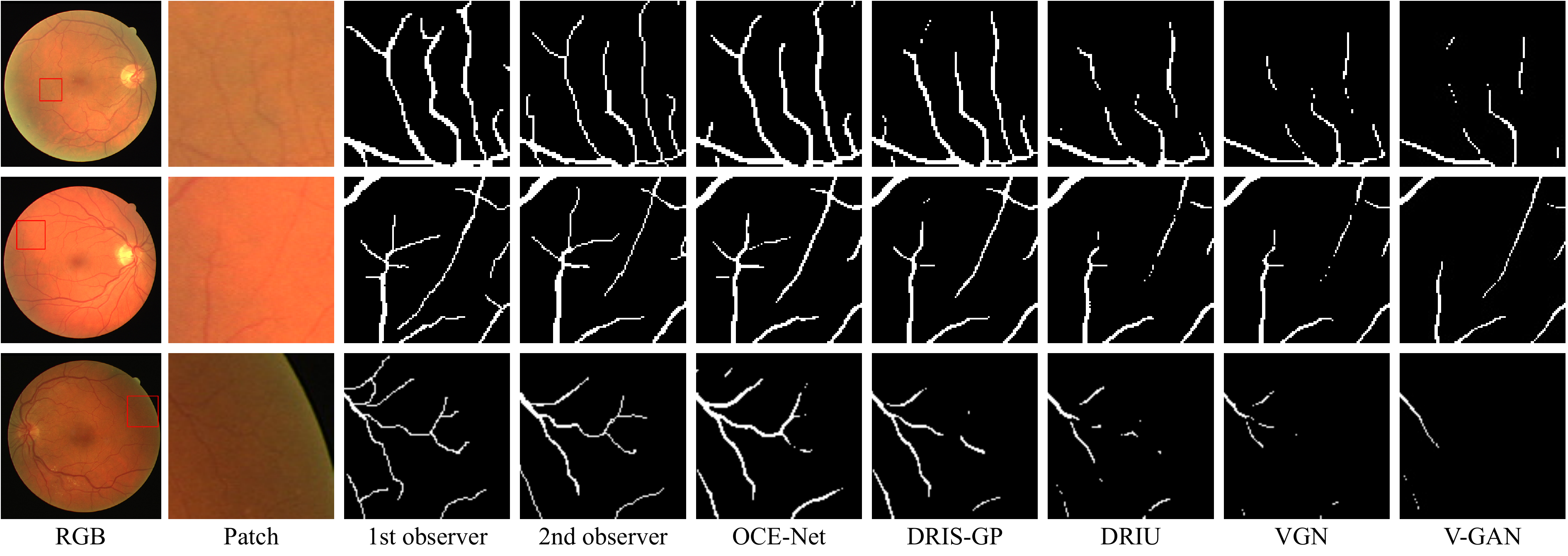}
	\caption{Visual comparison with other state-of-the-art methods for segmenting thin vessels.}
	\label{ex2}
\end{figure*}

Note that in Table. \ref{chasedb}, on the CHASEDB1 dataset, CTF-Net \cite{wang2020ctf} achieved better F1 score and SP than our OCE-Net, which is mainly because the fundus images in CHASEDB1 are quite different from the images in the DRIVE and STARE datasets. As shown in Fig. \ref{ex2}, there are almost no thin vessels in the images from CHASEDB1, but containing mainly thick vessels. In addition, the orientations of the blood vessels in CHASEDB1 are also less complicated. However, our newly designed modules for OCE-Net are mainly used to deal with thin vessels with complex orientations and contexts. Therefore, our OCE-Net performed less effective on CHASEDB1 than on DRIVE and STARE. 
In addition, in terms of some novel indicators in Table. \ref{drive2}, SkelCon \cite{tan2022retinal} achieved better performance on Connectivity (C) and Consistency (L), because SkelCon specially designed modules to improve the connectivity and consistency of vessels.

\begin{figure*}[htbp]
	\centering
	\includegraphics[width=14cm]{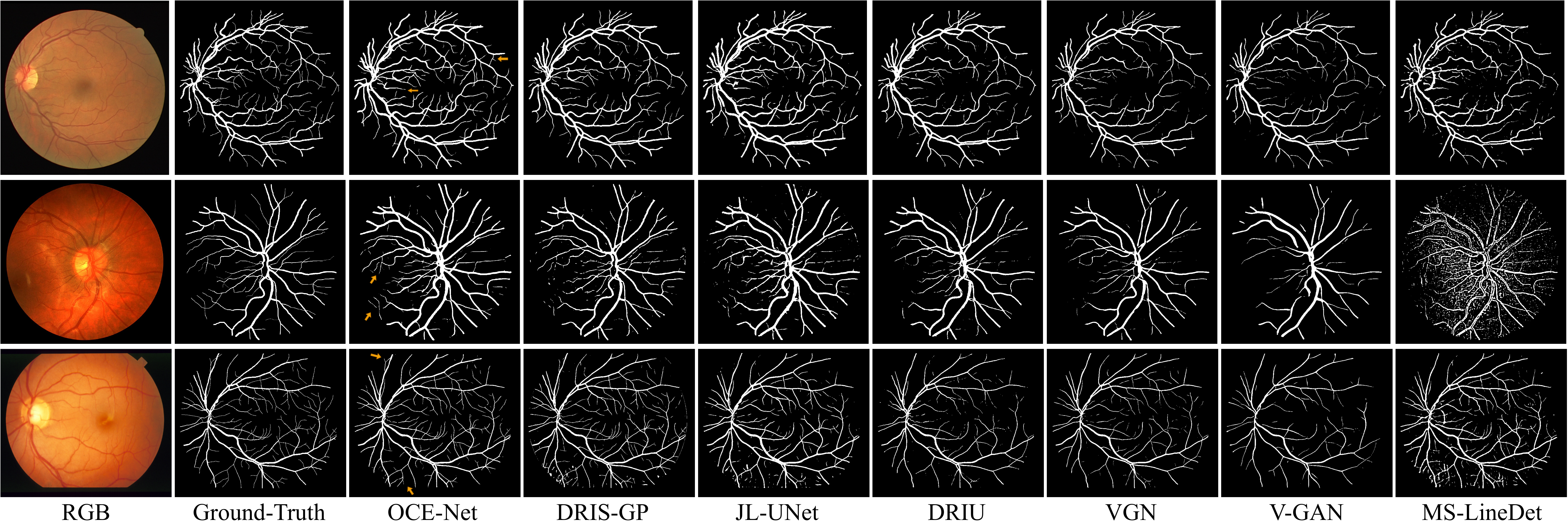}
	\caption{Visual comparison with other state-of-the-art methods on DRIVE, CHASEDB1 and STARE datasets from top to bottom rows. The orange arrows indicate some details of segmentation. Please zoom in for better view.}
	\label{ex1}
\end{figure*}

As shown in Fig. \ref{ex1}, on these three widely-used datasets, Our method achieves better visual segmentation results than the previous methods. Compared with other methods in Fig. \ref{ex2}, our method can effectively segment the thin blood vessels. In contrast, many other methods can not segment thin vessels well.

In addition, we also conducted quantitative comparison of detecting thin vessels on DRIVE. We used morphological image processing to separate out thin vessels. We chose three novel indicators including Connectivity (C), Overlapping Area (A) and Consistency (L), proposed in \cite{yan2017skeletal}, because thin vessels are usually very small, therefore, it is not suitable to evaluate them with ACC and AUC. As shown in Table. \ref{thin}, our OCE-Net achieved better performance on detecting thin vessels than other methods.

\begin{table}[]
	\footnotesize
	\renewcommand\arraystretch{1.1}
	\centering
	\caption{Experiments conducted on DRIVE for performance comparison of segmenting thin vessels.}
	\begin{tabular}{c|ccc}
		\toprule
		Method                     & Connectivity (C)    & Overlapping Aera (A)     & Consistency (L)      \\
		\midrule
		UNet           &91.34   & 85.72  &78.66  \\
		Attention UNet 
		 &92.03       &87.89       &80.14           \\
		 
		 Dense UNet 
		 &92.27       &88.02       &80.36           \\

		\midrule
		\textbf{OCE-Net}    &\color{red}92.45  &\color{red}88.23   &\color{red}80.68       \\ 
		\bottomrule
	\end{tabular}
	\label{thin}
\end{table}


\begin{table}[htbp]
	\footnotesize
	\renewcommand\arraystretch{1.1}
	\centering
	\caption{Quantitative comparison with other state-of-the-art methods on \textbf{DRIVE}. \color{red}Red: the best, \color{blue}Blue: the second best.}
	\begin{tabular}{c|c|ccccc}
		\toprule
		Method    &Year     & F1  & Se & Sp & Acc & AUC \\ \midrule

		2nd observer \cite{staal2004ridge}   &2004 &N.A  & 77.60 & 97.24 & 94.72 & N.A   \\

		HED \cite{xie2015holistically} &2016 & 80.89 & 76.27 & 98.01 & 95.24 & 97.58 \\

		DeepVessel \cite{fu2016deepvessel} &2016 & N.A & 76.12 & 97.68 & 95.23 & 97.52  \\

		Orlando et al. \cite{orlando2016discriminatively}  &2017                                                   &N.A    &78.97    &96.84     &94.54     &95.06  \\

		JL-UNet \cite{yan2018joint}  &2018                                                    &81.02    &76.53    &98.18     &95.42     &97.52   \\

		CC-Net \cite{feng2020ccnet}  &2018                                                    &N.A    &76.25    &98.09     &95.28     &96.78   \\

		Att UNet \cite{oktay2018attention} &2018                                                     
		&82.32  &79.46    &97.89    &95.64     &97.99   \\

		Dense UNet \cite{li2018h} &2018                                                     
		&\color{blue}82.79  &79.85    &98.05    &95.73     &98.10   \\ 
		
		Yan et al. \cite{yan2018three} &2019                                                   &N.A     &76.31    &\color{blue}98.20    &95.33     &97.50   \\

		BTS-DSN \cite{guo2019bts} &2019                                                     &82.08     &78.00    &98.06    &95.51     &97.96    \\ 
		
		DUNet \cite{jin2019dunet} &2019                                                     &82.49     &79.84    &98.03    &\color{blue}95.75     &\color{blue}98.11   \\  
		
		CTF-Net \cite{wang2020ctf} &2020                                                    &82.41    &78.49    &98.13     &95.67     &97.88   \\ 
		
		CSU-Net \cite{wang2020csu}  &2021                                                    &82.51    &\color{red}80.71    &97.82     &95.65     &98.01   \\ \midrule
		\textbf{OCE-Net (Ours)}   &2022   &\color{red}83.02   &\color{blue}80.18   &\color{red}98.26  &\color{red}95.81  &\color{red}98.21 \\  \bottomrule

	\end{tabular}
	\label{drive}
\end{table}

\begin{table}[htbp]
	\footnotesize
	\renewcommand\arraystretch{1.1}
	\centering
	\caption{Quantitative comparison with other state-of-the-art methods on \textbf{STARE} dataset.}
	\begin{tabular}{c|c|ccccc}
		\toprule
		Method    &Year     & F1  & Se & Sp & Acc & AUC \\ \midrule


		HED \cite{xie2015holistically}  &2016  & 82.68 &\color{blue} 80.76 & 98.22 & 96.41 & 98.24 \\


		Orlando et al. \cite{orlando2016discriminatively}  &2017                                                   &N.A    &76.80    &97.38     &95.19     &95.70  \\

		JL-UNet \cite{yan2018joint} &2018                                                    &N.A    &75.81    &98.46     &96.12     &98.01   \\

		Att UNet \cite{oktay2018attention} &2018                                                     
		&81.36     &80.67    &98.16    &96.32     &98.33   \\

		CC-Net \cite{feng2020ccnet} &2018                                                     
		&N.A     &77.09    &98.48    &96.33     &97.00   \\

		Dense UNet \cite{li2018h} &2018                                                     
		&82.32     &78.59    &98.42    &96.44     &98.47   \\ 
		
		Yan et al. \cite{yan2018three} &2019                                                   &N.A     &77.35    &\color{blue}98.57    &96.38     &98.33   \\

		BTS-DSN \cite{guo2019bts} &2019                                                     &\color{red}83.62     &\color{red}82.01    &98.28    &\color{blue}96.60     &\color{blue}98.72    \\ 
		
		DUNet \cite{jin2019dunet} &2019                                                     &82.30     &78.92    &98.16    &96.34     &98.43   \\  \midrule
		
		\textbf{OCE-Net (Ours)} &2022  &\color{blue}83.41 &80.12 &\color{red}98.65 &\color{red}96.72  &\color{red}98.76    \\
		
		\bottomrule

	\end{tabular}
	\label{stare}
\end{table}

\begin{table}[htbp]
	\footnotesize
	\renewcommand\arraystretch{1.1}
	\centering
	\caption{Quantitative comparison with other state-of-the-art methods on \textbf{CHASEDB1} dataset.}
	\begin{tabular}{c|c|ccccc}
		\toprule
		Method    &Year     & F1  & Se & Sp & Acc & AUC \\ \midrule

		2nd observer \cite{staal2004ridge}   &2004 &N.A  & \color{blue}81.05 & 97.11 & 95.45 & N.A   \\

		HED \cite{xie2015holistically} &2016  & 78.15 & 75.16 & 98.05 & 95.97 & 97.96 \\

		DeepVessel \cite{fu2016deepvessel} &2016 & N.A & 74.12 & 97.01 & 96.09 & 97.90  \\

		Orlando et al. \cite{orlando2016discriminatively}  &2017                                                   &N.A     &75.65     &96.55      &94.67    &94.78  \\

		JL-UNet \cite{yan2018joint}  &2018                                                    &N.A     &76.33     &98.09      &96.10     &97.81   \\

		Att UNet \cite{oktay2018attention} &2018                                                     
		&80.12     &80.10     &98.04     &96.42      &98.40   \\

		Dense UNet \cite{li2018h} &2018                                                     
		&79.01     &78.93     &97.92     &96.11      &98.35   \\ 
		
		Yan et al. \cite{yan2018three} &2019                                                   &N.A     &76.41     &98.06     &96.07      &97.76   \\

		BTS-DSN \cite{guo2019bts} &2019                                                     &79.83     &78.88     &98.01     &96.27      &98.40    \\ 
		
		DUNet \cite{jin2019dunet} &2019                                                     &79.32     &77.35     &98.01     &96.18      &98.39   \\ 
		CTF-Net \cite{wang2020ctf} &2019                                                     &\color{red}82.20     &79.48     &\color{red}98.42     &\color{blue}96.48      &\color{blue}98.47   \\

		\midrule
		
		
		\textbf{OCE-Net (Ours)} &2022  &\color{blue}81.96  &\color{red}81.38   &\color{blue}98.24   &\color{red}96.78    &\color{red}98.72    \\  \bottomrule

	\end{tabular}
	\label{chasedb}
\end{table}

\begin{table}[]
	\footnotesize
	\renewcommand\arraystretch{1.2}
	\setlength\tabcolsep{3pt}
	\centering
	\caption{Quantitative comparison of several newly proposed metrics \cite{yan2017skeletal} between our method
		and other methods on \textbf{DRIVE} dataset. Note that the model of SA-UNet \cite{guo2021sa} was reproduced by us in order to eliminate the inconsistency of indicators in different papers \cite{jin2019dunet} \cite{guo2021sa}.}
	\begin{tabular}{c|cccccccc}
		\toprule
		Method               & F    & C    & A    & L   & rSe  &rSp  &rAcc  &Mcc \\
		\midrule
		2nd observer        & 83.75 & 100 & 93.98 & 89.06 & 85.84  &99.19   &95.74   &76.00 \\ \midrule
		
		HED \cite{xie2015holistically}       & 80.09 & 99.75 & 90.06 & 89.11 & 71.57 &95.11   &89.08   &66.00 \\
		DRIU \cite{maninis2016deep}       & 80.43 & 99.56 & 91.52 & 88.23 & 82.36  &96.85   &93.13  &71.61 \\
		DeepVessel \cite{fu2016deepvessel}       & 61.74 & 99.60 & 84.23 & 73.38 & 54.93  &99.78   &88.32   &73.34 \\
		V-GAN \cite{son2017retinal}       & 84.82 & 99.64 & 94.69 & 89.84 & 80.77  &\color{blue}99.63   &94.76   &80.24 \\
		
		JL-UNet \cite{yan2018joint}       & 81.06 & 99.61 & 93.08 & 87.35 & 76.11  &99.57   &93.53   &78.98 \\
		SWT-FCN \cite{oliveira2018retinal}        &83.92       &99.73       &94.36       &89.11       &79.63     &99.64   &94.48   &\color{blue}80.53    \\
		DeepDyn \cite{khanal2020dynamic}          &84.53       &90.70       &94.58       &89.61       &81.52   &99.44  &\color{blue}94.82  &80.02       \\
		DAP \cite{sun2021robust}      &82.55       &99.72       &93.74       &88.24       &78.57   &99.57    &94.15    &79.00       \\
		DRIS-GP \cite{cherukuri2019deep}          &84.94      &99.68       &\color{blue}94.91       &89.74       &80.22  &\color{red}99.64   &94.66   &\color{red}81.84     \\
		SA-UNet \cite{guo2021sa}       & 83.19 & 99.61 & 93.96 & 88.89 & 80.01 &99.32   &94.53   &79.67 \\
		
		SkelCon \cite{tan2022retinal}      &\color{blue}85.30       &\color{red}99.85       &93.58       &\color{red}91.26      &\color{blue}83.23   &98.59    &94.61    &80.30       \\
		\midrule

		\textbf{OCE-Net}  &\color{red}85.83       &\color{blue}99.80       &\color{red}95.07       &\color{blue}90.43       &\color{red}83.83  &99.29   &\color{red}95.30   &80.40  \\  \bottomrule
		
	\end{tabular}
	\label{drive2}
\end{table}

\subsection{Comparison and ablation study of individual modules}

\subsubsection{Comparison between the proposed GLFM and other attention gates}
The proposed GLFM can be viewed as an alternative of attention gates of U-shaped networks. We compared our GLFM with other attention blocks and self-attention blocks by individually adding different attention gates to the baseline UNet. Among the methods listed in Table. \ref{ex_glfm}, the original atention gate in \cite{oktay2018attention} and CBAM block \cite{woo2018cbam} can be viewed as the modules that only carry out 'local' attention. The NL, DNL, self-attention blocks can be seen as the methods that only apply 'global' attention. In contrast, the proposed GLFM not only exerts 'local' attention, but also applies 'global' attention. 

From Table. \ref{ex_glfm}, the proposed GLFM outperforms all the 'local'-only and 'global'-only attention modules. This demonstrates that it is important to capture both of the global context information and the local details of vessels, because global context represents the overall structure information of blood vessels, while the local details focus more on the thin vessels. Without local attention, thin vessels would be missed easily. 

\begin{table}[]
	\footnotesize
	\renewcommand\arraystretch{1.1}
	\centering
	\caption{Comparison between the proposed GLFM and other prevalent modules serving as attention gates on DRIVE.}
	\begin{tabular}{c|ccccc}
		\toprule
		Method                     & F1    & Se    & Sp    & Acc   & AUC   \\
		\midrule
		Baseline (UNet) \cite{ronneberger2015u}            & 82.11 & 79.48 & 97.94 & 95.59 & 97.85 \\
		+ Attention Gate \cite{oktay2018attention}  &82.33       &79.12       &\color{blue}98.09       &95.68       &98.02       \\
		+ CBAM \cite{woo2018cbam}            &82.21       &79.26       &98.01       &95.64       &97.92       \\
		+ Non-local \cite{wang2018non}       &82.57       &\color{blue}79.88       &98.04       &95.73       &98.10       \\
		+ DNL \cite{yin2020disentangled}             &82.37       &78.99       &98.02       &95.72       &98.10       \\
		+ Self-Attention \cite{vaswani2017attention}  &\color{blue}82.66       &79.32       &\color{red}98.12       &\color{blue}95.73       &\color{blue}98.12       \\
		\midrule
		+ \textbf{GLFM (Proposed)} &\color{red}83.00       &\color{red}80.90       &97.92       &\color{red}95.76       &\color{red}98.16    \\ 
		\bottomrule
	\end{tabular}
	\label{ex_glfm}
\end{table}

\subsubsection{Comparison between the proposed OCE-NL/OCE-DNL and other modules serving as core fusion module in MSFM}
The proposed OCE-NL and OCE-DNL act as the core modules of the MSFM, which are used to entangle the context and orientation information. In order to prove the superiority of this entanglement mechanism, we conducted comparison by directly replacing the OCE-NL (OCE-DNL) with other prevalent modules for fair comparison, including vanilla NL, DNL and self-attention modules. The results shown in Table. \ref{ex_ocenl} demonstrate that entangling orientation features with the output features of the network can effectively improve the reconstruction of vessels compared with vanilla NL and DNL. Because orientation features provide more concentration on blood vessels via the oriented constraints of DCOA Conv, which can act as a kind of auxiliary prior information and help the network reconstruct vessels better. 

In Table. \ref{ex_ocenl}, 'Addition + DNL' means the plain and orientation features are added together by element-wise addition operation and then fed into DNL for fusion. 'Concat + DNL' means the plain and orientation features are concatenated together and a Conv1x1 operator is then applied to the features for dimension reduction, and then fed into DNL for fusion. The addition, concatenation and our proposed cross-correlated entanglement can be regarded as three independent ways for feature fusion. Compared with DNL without using orientation prior features, both 'Addition + DNL' and 'Concat + DNL' (both the plain and orientation features involved) gain a significant improvement over the 'plain features only' method, which indicates that orientation features can indeed help segment vessels better. Among the 'both plain and orientation features involved' methods, our proposed entanglement is the best way to integrate the plain and orientation features together.

\begin{table}[htbp]
	\footnotesize
	\renewcommand\arraystretch{1.1}
	\centering
	\caption{Comparison between the proposed OCE-NL/OCE-DNL and other modules serving as core fusion block in MSFM on STARE.}
	\begin{tabular}{c|ccccc}
		\toprule
		Method                     & F1    & Se    & Sp    & Acc   & AUC   \\
		\midrule
		Baseline (UNet) \cite{ronneberger2015u}            & 80.87 & 74.82 & 98.78 & 96.24 & 98.17 \\
		+ Self-Attention \cite{vaswani2017attention}       &82.22       &\color{red}77.77       &98.64       &96.43       &98.40       \\
		+ Non-local (NL) \cite{wang2018non}            &81.06       &75.55       &98.87       &96.40       &98.46       \\
		+ DNL \cite{yin2020disentangled}  &80.89       &75.36       &98.96       &96.36       &98.49       \\
		\midrule
		Addition + DNL  &82.06       &\color{blue}76.21       &98.86       &96.46       &98.51       \\
		Concat + DNL  &\color{blue}82.32      &75.29      &98.89       &\color{blue}96.54       &98.53       \\
		\midrule
		+ \textbf{OCE-NL (Proposed)} &81.95       &75.17      &\color{blue}99.02       &96.49       &\color{blue}98.57  \\
		+ \textbf{OCE-DNL (Proposed)} &\color{red}82.37       &75.65       &\color{red}99.05       &\color{red}96.56       &\color{red}98.58 \\  \bottomrule
	\end{tabular}
	\label{ex_ocenl}
\end{table}

\subsubsection{Comparison between the proposed DCOA Conv and other convolutions}
We compared the proposed DCOA Conv with other prevalent variants of convolution. Note that we directly replaced all the vanilla convolution in the UNet with different convolution variants during testing for fair comparison. 

In Table. \ref{ex_dcoa}, 'Gabor Conv (4)/(8)' and 'DCOA Conv (4)/(8)' mean that filters with 4 or 8 different orientations were adopted. 'Dynamic Conv (4)/(8)' means 4 or 8 kernels were used. 

As shown in Table. \ref{ex_dcoa}, the proposed DCOA Conv outperforms other convolution operators when the number of orientation was set to 8. Note that Gabor Conv has side effect on vessel segmentation when it was inserted into the UNet baseline, because Gabor Conv can only encode a single orientation per channel, which can not capture the complex vessels with various orientations. This restriction limits the network to learn the complex orientation characteristics of blood vessels, resulting in poor performance. 

Dynamic Conv (4) and Dynamic Conv (8) have similar effects because Dynamic Conv improves the performance by adding more kernels, however, these kernels do not have orientation selectivity. When the number of kernels is 4 in Dynamic Conv, the feature extraction capability of network has reached its peak, therefore, increasing the number of kernels to 8 cannot improve the feature extraction capability. 

In comparison, the proposed DCOA Conv has the ability of capturing complex multiple orientations, which succeeds in overcoming the disadvantage of Gabor Conv. Through our experiments, eight orientations are enough for our model to work well on the DRIVE dataset, and adding more orientations would not contribute much. Especially, the vascular orientation in the CHASEDB1 dataset is much simpler than that in the DRIVE dataset, with fewer thin vessels and less complexity of orientations, therefore, four orientations can well encode the orientation information for the CHASEDB1 dataset.

Deformable Conv improves the performance by fitting the kernel's shapes to vessels (by learning the varying shapes of vessels). In comparison, our proposed DCOA Conv learns different orientations of vessels by integrating multiple oriented kernels together. These two methods improve the accuracy of vascular segmentation from two different starting points.

\begin{table}[htbp]
	\footnotesize
	\renewcommand\arraystretch{1.1}
	\centering
	\caption{Comparison and abaltion study between the proposed DCOA Conv and other prevalent convolution operators on DRIVE.}
	\begin{tabular}{c|ccccc}
		\toprule
		Method                     & F1    & Se    & Sp    & Acc   & AUC   \\
		\midrule
		Baseline (UNet) \cite{ronneberger2015u}           & 82.11 & 79.48 & 97.94 & 95.59 & 97.85 \\
		+ Gabor Conv (4) \cite{luan2018gabor}  &81.43       &78.85       &97.89       &95.48       &97.73       \\
		+ Gabor Conv (8) \cite{luan2018gabor}  &81.78       &77.99       &\color{red}98.12       &95.52       &97.82       \\
		+ DR Conv \cite{chen2021dynamic}           &82.11       &79.29       &98.03       &95.63       &97.93       \\
		+ Cond Conv \cite{yang2019condconv}           &81.94      &80.24       &97.86       &95.62       &97.94       \\
		+ Dynamic Conv (4) \cite{chen2020dynamic}       &82.22       &80.15       &98.00       &95.69       &98.08       \\
		+ Dynamic Conv (8) \cite{chen2020dynamic}       &\color{blue}82.31       &79.55       &\color{blue}98.03       &95.69       &98.08       \\
		+ Deformable Conv V1 \cite{dai2017deformable}            &82.42       &\color{blue}80.42       &97.93       &\color{blue}95.70       &\color{blue}98.11       \\
		+ Deformable Conv V2 \cite{zhu2019deformable}  &82.26       &79.82       &98.00       &95.69       &98.10       \\
		\midrule
		+ \textbf{DCOA Conv (4)} &82.28       &79.67       & 98.01      &95.64       &98.03 \\
		+ \textbf{DCOA Conv (8)} &\color{red}82.69       &\color{red}80.50       & 97.97      &\color{red}95.74       &\color{red}98.13 \\ \bottomrule 
	\end{tabular}
	\label{ex_dcoa}
\end{table}

\subsubsection{Why do we must fuse the plain and the orientation features together via SAFM?}
\label{subsec:534}

As shown in Fig. \ref{model}, we fused the plain features extracted by basic blocks and the orientation features extracted by DCOA blocks together in the network via SAFM. Why is this fusion essential? In order to answer this important question, we conducted experiments to demonstrate that the fusion is essential to involve the orientation information into the network. 

As shown in Table. \ref{ex_safm}, when we removed the DCOA block and SAFM from the OCE-Net, corresponding to the 'plain only' mode, which can not capture the orientation information of vessels well. When we directly replaced the plain conv with the proposed DCOA Conv, corresponding to the 'orientation only' mode, which only uses the DCOA Conv to extract orientation features. As we can see in Table. \ref{ex_safm}, the results of 'orientation only' mode are significantly lower than those of 'plain only' mode, which demonstrates that segmenating vessels with only orientation features would fail to gain promising results and even has side effects on the context-aware modules (GLFM, OCE-DNL), because except for the orientation features, other features such as fundus background and other tissues, can serve as negtive samples against the blood vessels (positive samples) and also play an important role in accurately detecting vessels. Therefore, orientation features should be regarded as a kind of auxiliary prior guidance, used to help the plain features for better reconstruction of vessels.

Considering that there are some redundant channels in both features and there is correlation between the channels of the two features, direct fusion using 1x1 convolution cannot achieve the best performance, what's worse, when we replaced SAFM with Conv1x1 (the most commonly used fusion module) as the fusion module, the segmentation performance declines seriously, as indicated by the underlined results in Table. \ref{ex_safm}. In comparison, adopting SAFM as fusion module in the network instead of Conv1x1 can gain a distinct improvement.

\begin{table}[]
	\footnotesize
	\renewcommand\arraystretch{1.1}
	\centering
	\caption{Experiments conducted on CHASEDB1 for explaining for the necessity of SAFM. The \underline{underlined results} indicate significant declined performance.}
	\begin{tabular}{c|ccccc}
		\toprule
		Method                     & F1    & Se    & Sp    & Acc   & AUC   \\
		\midrule
		Plain only           &\color{blue}81.42 & \color{red}83.19   &97.97 & \color{blue}96.56 & \color{blue}98.62 \\
		Orientation only 
		(DCOA)  &\underline{81.01}       &\color{blue}82.74       &\underline{97.84}       &\underline{96.47}       &\underline{98.48}       \\
		Fusion by Conv1x1         &\underline{71.76}       &\underline{60.76}       &\color{red}98.96       &\underline{94.97}      &98.61       \\
		\midrule
		\textbf{Fusion by SAFM}    &\color{red}81.59  &82.25   &\color{blue}98.11   &\color{red}96.66    &\color{red}98.67      \\ 
		\bottomrule
	\end{tabular}
	\label{ex_safm}
\end{table}

\subsubsection{Comparison between the proposed UARM and the vanilla spatial attention as the refining module}
\begin{figure}[htbp]
	\centering
	\includegraphics[width=11cm]{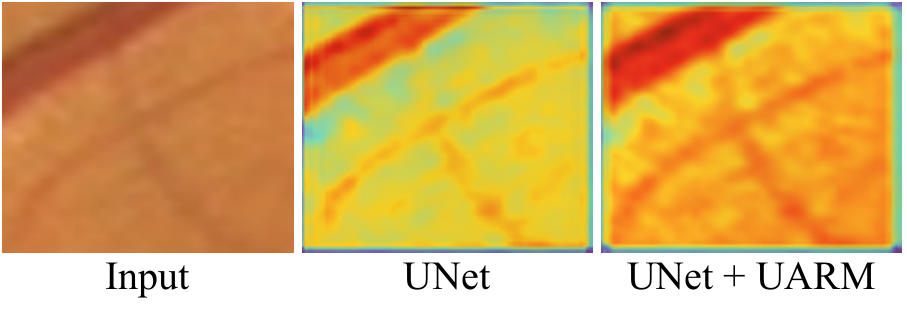}
	\caption{Ablation study of UARM in terms of visual attention map. More attention that network is paid on the thin vessels.}
	\label{abla_uarm}
\end{figure}

At the end of the network, the proposed UARM is adopted to refine the final output features by dealing with the unbalance problem among the fundus background, thick and thin vessels. The vanilla spatial attention is used here as the competitor in order to prove the effectiveness of our UARM. As shown in Table. \ref{ex_uarm}, the vanilla spatial attention gains an improvement, and the proposed UARM is better than the vanilla spatial attention. In the UNet baseline, the attention of the network focuses more on the thick vessels which have more salient features and the thin vessels are usually neglected, resulting in low segmentation accuracy. As shown in Fig. \ref{abla_uarm}, UARM can allocate more attention to distinguish thin vessels and obtain a significant promotion.

\begin{table}[htbp]
	\footnotesize
	\renewcommand\arraystretch{1.1}
	\centering
	\caption{Comparison between the proposed UARM and the vanilla spatial attention serving as the final refining module on DRIVE.}
	\begin{tabular}{c|ccccc}
		\toprule
		Method                     & F1    & Se    & Sp    & Acc   & AUC   \\
		\midrule
		Baseline (UNet) \cite{ronneberger2015u}           & 82.11 & \color{blue}79.48 & 97.94 & 95.59 & 97.85 \\
		+ Spatial Attention \cite{qin2020ffa}  &\color{blue}82.19       &79.47       &\color{blue}98.01       &\color{blue}95.66       &\color{blue}98.04       \\
		\midrule
		+ \textbf{UARM (Proposed)} &\color{red}82.61       &\color{red}80.02       &\color{red}98.03       &\color{red}95.73       &\color{red}98.10  \\  
		\bottomrule
	\end{tabular}
	\label{ex_uarm}
\end{table}

\begin{figure*}[htbp]
	\centering
	\includegraphics[width=14cm]{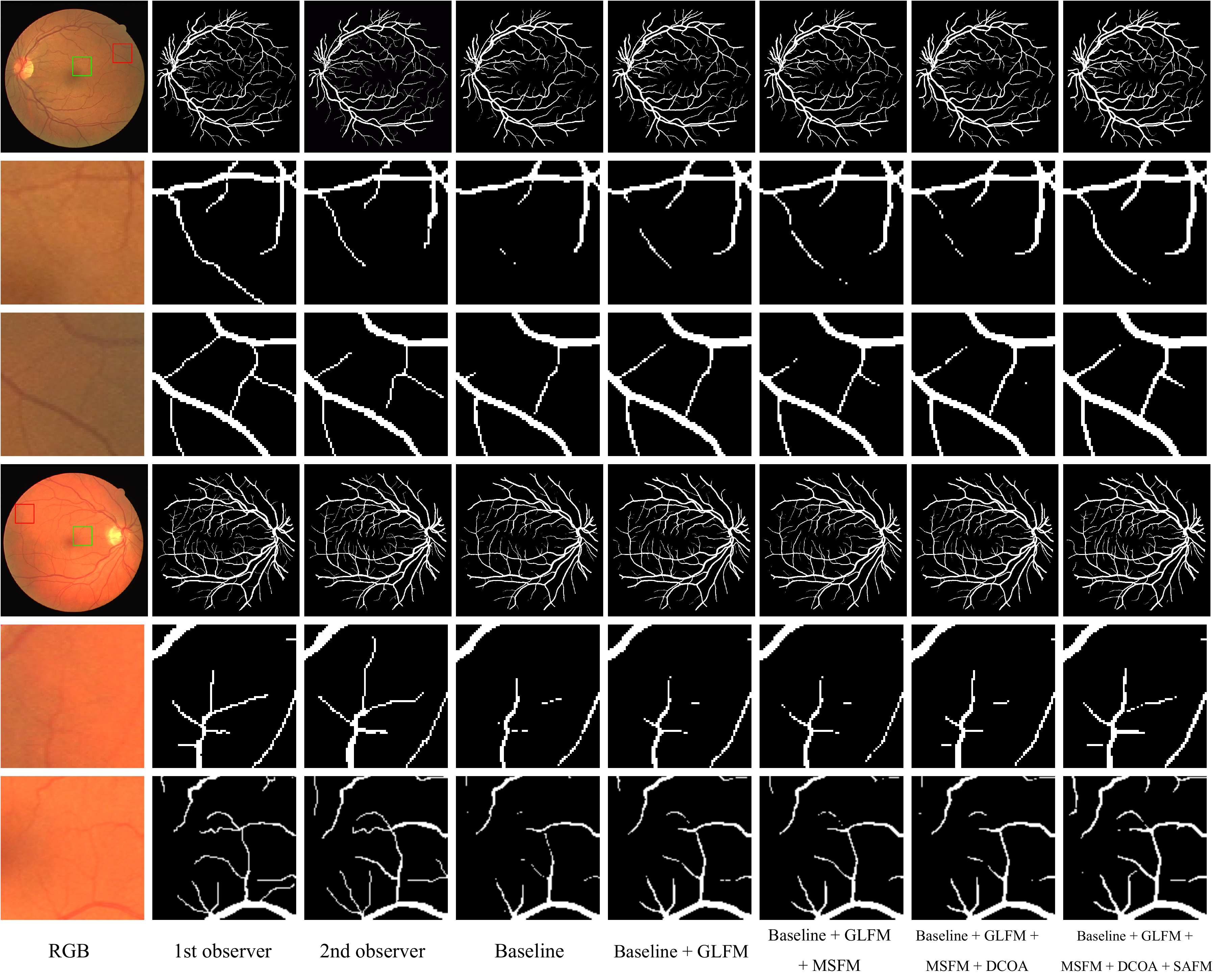}
	\caption{Visual ablation study of each proposed module. Notice the last second column, when we directly replaced plain conv with DCOA Conv, the visual effect got worse, however, as shown in the last column, when we used SAFM to fuse the plain and orientation features, an improvement was witnessed. The corresponding quantitative results are listed in Table. \ref{abla}.}
	\label{ex3}
\end{figure*}

\begin{table}[htbp]
	\scriptsize
	\renewcommand\arraystretch{1.15}
	\setlength\tabcolsep{3pt}
	\centering
	\caption{Overall ablation study for each proposed module on CHASEDB1. The \underline{underlined results} indicate significant declined performance.}
	\begin{tabular}{l|ccccc}
		\toprule
		\qquad \qquad \qquad \qquad \qquad  Method                     & F1    & Se    & Sp    & Acc   & AUC   \\
		\midrule
		Baseline (UNet)            & 79.22 & 79.10 & 97.81 & 96.30 & 98.22 \\
		\midrule
		+ DCOA Conv  & 80.71       &79.38       &98.11       &96.45  &98.47  \\
		+ GLFM  &80.61       &79.93       &98.09       &96.47       &98.48       \\
		+ GLFM + OCE-DNL            &81.33       &82.03       &97.98       &96.54       &98.58       \\
		\midrule
		+ GLFM + OCE-DNL + DCOA Conv       &\underline{81.01}       &\underline{81.24}       &\underline{97.92}       &\underline{96.50}       &\underline{98.48}       \\
		+ GLFM + OCE-DNL + DCOA Conv + SAFM             &81.51       &82.17       &98.08       &96.61       &98.63       \\
		\midrule
		+ GLFM + OCE-DNL + Deformable Conv       &\underline{80.31}       &\underline{79.41}       &98.32       &\underline{96.51}       &\underline{98.49}       \\
		+ GLFM + OCE-DNL + Deformable Conv + SAFM             &81.29       &81.86       &98.12       &96.60       &98.62       \\
		\midrule
		+ GLFM + OCE-DNL + DCOA Conv + SAFM + UARM    &\textbf{81.59}  &\textbf{82.25}   &\textbf{98.11}   &\textbf{96.66}    &\textbf{98.67}       \\
		\bottomrule
	\end{tabular}
	\label{abla}
\end{table}

\subsection{Overall ablation study for each proposed modules}
In Table. \ref{abla}, we conducted an overall ablation study for each of the proposed modules on CHASEDB1. Note that directly replacing the plain convolution with DCOA Conv would make the model worse and yield a dramatic performance reduction, as indicated by the underlined results in Table \ref{abla}. Only with SAFM can plain features and orientation features be well fused. Note that the model referred here is the model equipped with the context-aware modules, such as GLFM and OCE-DNL. Directly replacing the plain conv with the DCOA Conv in the vanilla UNet can actually gain an improvement, which indicates that there is an unexpected conflict between the vessel-aware conv and context-aware modules. This conflict will be discussed later in Section. \ref{conflict}.

It is worth mentioning that directly replacing the plain convolution with Deformable Conv \cite{dai2017deformable} \cite{zhu2019deformable} would unexpectedly make the performance (of the model with context-aware modules) worse as well. The reason is similar with the DCOA Conv mentioned before, because the Deformable Conv reshapes the convolution kernels (DCOA Conv also changes the shape of convolution kernels) to fit the vessels and inevitably neglects other important features like fundus background and other tissues, which can serve as important negative samples to help better detect blood vessels. When we adopted SAFM to fuse the plain features and the features extracted by deformable convolution together, a promotion on segmentation was also obtained like that when we fused the plain and orientation features together before, which emphasizes again that it is essential to integrate the plain features and the vessel-aware features (shape-aware or orientation-aware) together.

\subsection{The conflict between the vessel-aware conv and the context-aware modules}
\label{conflict}

We reclaim that drictly inserting DCOA Conv or Deformable Conv into a vanilla UNet without the context-aware modules (GLFM, OCE-DNL) can actually gain an improvement, but the performance of context-aware modules will be degraded when they work together with vessel-aware convolution. In other words, there is a conflict between the vessel-aware conv and the context-aware modules. The reasons are as follows.
The vessel-aware conv extracts the local features with special orientations (DCOA Conv) or shapes (Deformable Conv), however, other features (background and non-vascular tissues) can not be extracted and encoded in these vessel-aware features. In contrast, context-aware modules work to capture the global context information by computing all the features in the fundus images, including the vascular and non-vascular features, which is the reason why 'orientation-only' mode in Table. \ref{ex_safm} gains a relatively worse performance. This demonstrates again that both the plain and the orientation features are equally important.
The function of the proposed SAFM is exactly to tackle the conflict and make the vessel-aware conv compatible with those context-aware modules, and our OCE-DNL is also a novel approach to entangle the vessel-aware features into the context information.

\begin{figure*}[htbp]
	\centering
	\includegraphics[width=15cm]{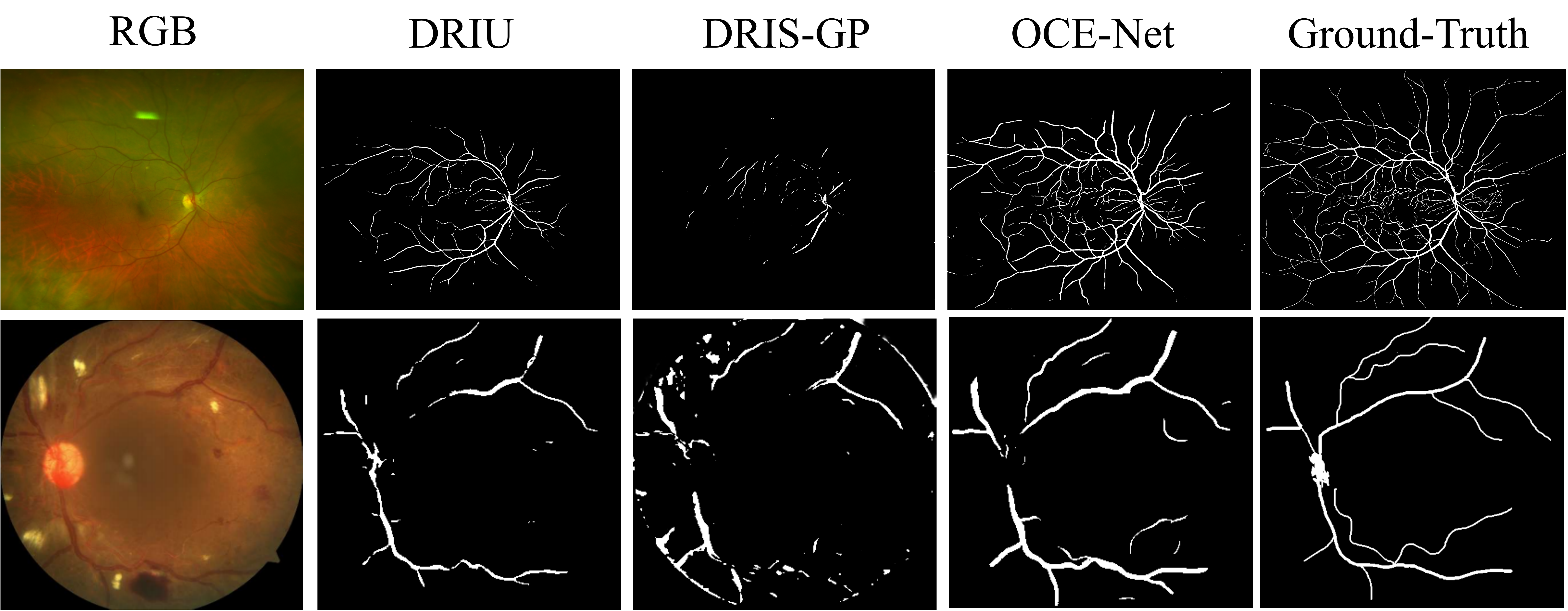}
	\caption{Visual comparison on the AV-WIDE (the first row) and UoA-DR (the second row) datasets.}
	\label{test2}
\end{figure*}

\begin{figure*}[htbp]
	\centering
	\includegraphics[width=15cm]{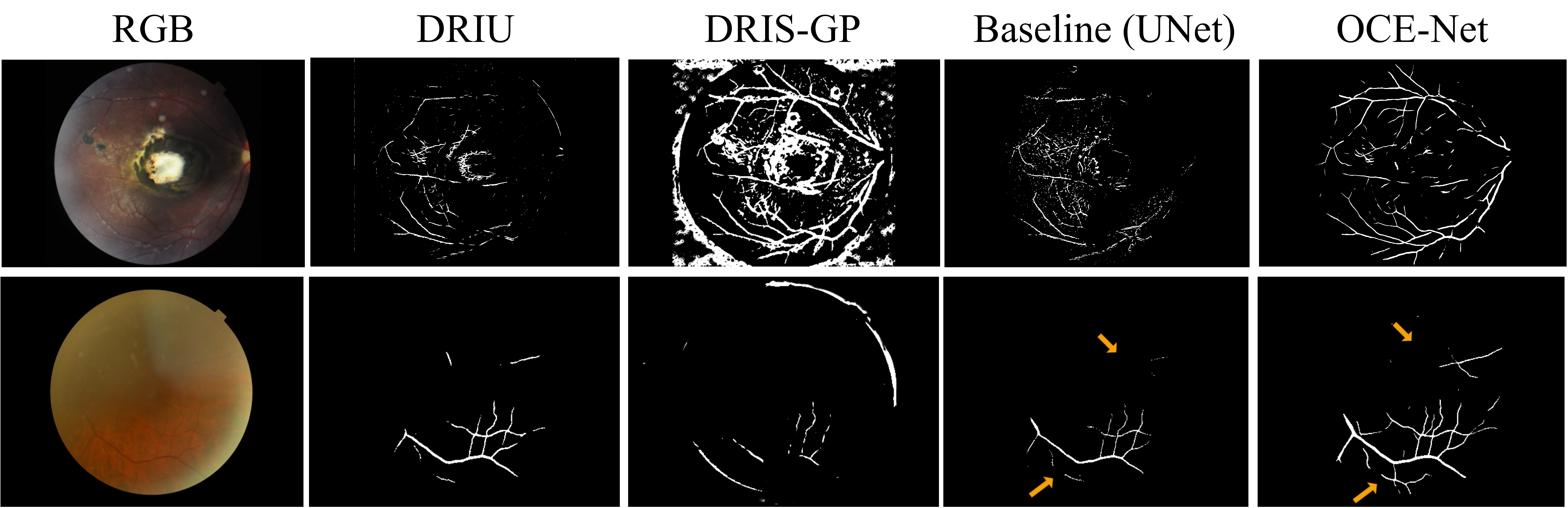}
	\caption{Visual comparison on the RFMiD (the first row) and UK Biobank (the second row) datasets.}
	\label{test1}
\end{figure*}

\subsection{Comparison with other methods on other challenging test sets}
In order to evaluate the generalization and robustness of our OCE-Net, we tested our model on several challenging test sets, including  AV-WIDE \cite{estrada2015retinal}, UoA-DR \cite{chalakkal2017comparative}, RFMiD \cite{pachade2021retinal} and UK Biobank \cite{sudlow2015uk}. Note that all the models involved in comparison were trained on DRIVE instead of re-training from scratch on these challenging datasets. For fair comparison, all the models involved here were trained and tested with grayscale fundus images rather than RGB images.

As shown in Fig. \ref{test2} and \ref{test1}, our method outperforms other methods, including DRIU \cite{maninis2016deep} and DRIS-GP \cite{cherukuri2019deep}, and gains a significant improvement over the baseline UNet \cite{ronneberger2015u}. Note that in the test on the UK Biobank \cite{sudlow2015uk} in the second row of Fig. \ref{test1}, the orange arrows indicate the area where the thin vessels are severely obscured by the opacity, which are difficult to be detected, even with human eyes. In contrast, our method can detect these blurred and occluded vessels well, which demonstrates the strong power of the proposed OCE-Net.

\begin{table}[htbp]
	\footnotesize
	\renewcommand\arraystretch{1.2}
	\setlength\tabcolsep{4pt}
	\centering
	\caption{Cross validation across the DRIVE and STARE datasets.}
	\begin{tabular}{c|c|cccc}
		\toprule
		Test set                                                                              & Method  & Se    & Sp    & Acc   & AUC   \\ \midrule
		\multirow{3}{*}{\begin{tabular}[c]{@{}c@{}}DRIVE \\ (trained on STARE)\end{tabular}} & CC-Net \cite{feng2020ccnet}  & 72.17 & \color{blue}98.20 & \color{blue}94.86 & 93.27 \\ 
		& BTS-DSN \cite{guo2019bts} & \color{blue}72.92 & 98.15 & \color{red}95.02 & \color{blue}97.09 \\ 
		& \textbf{OCE-Net} & \color{red}75.36 & \color{red}98.62 & 94.65 & \color{red}97.32 \\ \midrule
		\multirow{3}{*}{\begin{tabular}[c]{@{}c@{}}STARE \\ (trained on DRIVE)\end{tabular}} & CC-Net \cite{feng2020ccnet}  & \color{blue}74.99 & 97.98 & \color{red}95.63 & \color{blue}96.21 \\ 
		& BTS-DSN \cite{guo2019bts} & 71.88 & \color{blue}98.16 & 95.48 & 94.86 \\ 
		& \textbf{OCE-Net} & \color{red}75.06 & \color{red}98.28 & \color{blue}95.53 & \color{red}96.23 \\ \bottomrule
	\end{tabular}
	\label{cross}
\end{table}

\subsection{Cross validation}
In order to test the generalization ability, we conducted cross validation experiments on DRIVE and STARE and compared with other methods. 
As shown in Table. \ref{cross}, compared with other methods, our method achieves promising performance in terms of Se, Sp and AUC and comparable 
performance in terms of Acc. This demonstrates the relatively better 
generalization ability of the proposed model.

\begin{figure}[htbp]
	\centering
	\includegraphics[width=14cm]{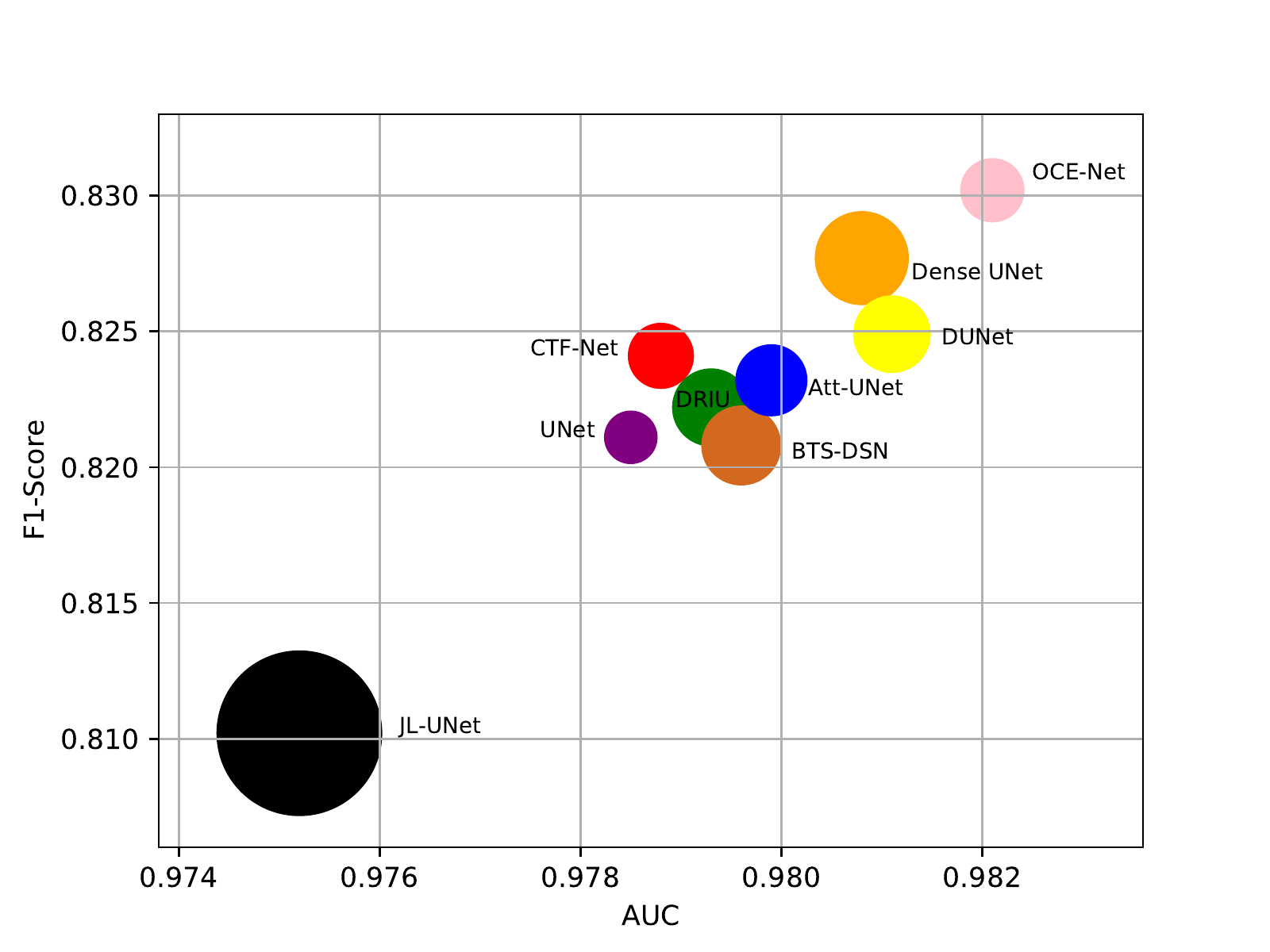}
	\caption{Comparison of model parameters. Note that the size of
		circle indicates the number of model parameters. F1 score and AUC are used to evaluate the performance of models.}
	\label{para}
\end{figure}

\subsection{Comparison of Parameters}
As shown in Fig. \ref{para}, we compared our OCE-Net with other state-of-the-art methods in terms of model parameters, F1 score and AUC. Our OCE-Net uses plain UNet as the backbone. The parameters of the UNet model are about 3.4MB. Our OCE-Net has about 6.3MB parameters after adding all the proposed modules (DCOA Conv, GLFM, OCE-DNL and UARM) into the UNet backbone. In contrast, JL-UNet \cite{yan2018joint} is a large model with about 33MB parameters, however, it achieved the worst performance. CTF-Net \cite{wang2020ctf}, DRIU \cite{maninis2016deep}, Attention UNet \cite{oktay2018attention} and BTS-DSN \cite{guo2019bts} have similar numbers of parameters (around 7MB) and similar performance. In addition, Dense Unet has about 11MB parameters and Deformable UNet (DUNet) has 7.4MB parameters, but both of them perform less effective than our OCE-Net. Our OCE-Net achieves the best performance in terms of F1 score and AUC without introducing too much parameters.

\section{Discussion}
\label{sec:discussion}
We compared our OCE-Net with many other state-of-the-art methods in terms of both quantitative and qualitative ways. We reproduced the network models of many UNet-based methods and retrained them from scratch on our codes, such as Attention UNet and Dense UNet. And some methods provide original codes and segmentaion results, such as JL-UNet and SkelCons, therefore we used their segmentation results for the comparison. However, some methods did not provide codes and results, such as DeepVessel, CC-Net, BTS-DSN and CSU-Net, we can only compare our methods with the data in their papers. In general, we believe that the comparison can prove the advantage of our work.

\section{Conclusion}
\label{sec:conclusion}

In this paper, a novel model called OCE-Net is proposed to simultaneously capture the orientation and context information of blood vessels as well as entangle them together. 
To this end, a novel convolution operator is designed to fit the vessels with complex orientations, and the experimental results demonstrate that feature extraction along more specific orientations can improve vascular continuity and connectivity. In addition, a global and local fusion module is constructed to leverage both of the context and detail information of vessels, because context information can help the network perceive the whole vascular skeleton and deal with occlusion well. Moreover, a novel entanglement mechanism is developed to entangle the context and the orientation information by introducing cross-correlation into vanilla non-local. Finally, to deal with the unbalance among fundus background, thick and thin vessels, a novel attention module is proposed to refine the results by allocating more attention to the regions where the vessels have quite low discriminability. 

Though promising performance have been obtained, there is still space for improving our model, considering that retinal vessel 
segmentation is quite challenging. For example, when fundus images contain various lesions, the continuity and connectivity of the extracted 
vessels are quite poor. As an imprtant future work, we plan to build a graph model to perceive the entire vascular skeleton to better capture the context information and improve vascular connectivity.


\section{Acknowledgment}
This work was supported by Key Area R\&D Program of Guangdong Province (2018B030338001), National Natural Science Foundation of China (62076055) and Medico-Engineering Cooperation Funds from UESTC 

(ZYGX2022YGRH013).

\bibliographystyle{unsrt}
\bibliography{ref}

%
%
%


%

\end{document}